\documentclass{article}

\usepackage[final]{corl_2020} % Uncomment for the camera-ready ``final'' version.

\usepackage{graphicx}
\usepackage{comment}
\usepackage{amsmath,amssymb} % define this before the line numbering.
\usepackage{color}

\usepackage{mathtools}
\usepackage{booktabs}
\usepackage{rotating,tabularx}
\newcolumntype{T}{>{\raggedleft\arraybackslash} X}

\usepackage{bbm}
\usepackage{multirow}
\usepackage{bbding}
\usepackage{subcaption}
\usepackage{soul,color}
\usepackage{makecell, caption}

\usepackage{lipsum, fancyvrb}

\title{DROGON: A Trajectory Prediction Model based on Intention-Conditioned Behavior Reasoning}

\author{
  Chiho Choi$^{1}$\thanks{Corresponding author.}\quad Srikanth Malla$^1$\quad Abhishek Patil$^{1}$\thanks{Work done while employed at Honda Research Institute USA. Now at Hilti Group.}\quad Joon Hee Choi$^{2\ast}$\\
  $^1$Honda Research Institute, USA\quad
  $^2$Sungkyunkwan University, Korea \\
  \texttt{\{cchoi, smalla\}@honda-ri.com\quad patilnabhi@gmail.com\quad jhchoi2019@skku.edu} \\
}

\begin{document}
\maketitle

%===============================================================================

\begin{abstract}
    We propose a Deep RObust Goal-Oriented trajectory prediction Network (DROGON) for accurate vehicle trajectory prediction by considering behavioral intentions of vehicles in traffic scenes. Our main insight is that %a relationship between intention and behavior of drivers can be reasoned from the observation of their interactions toward an environment. 
    the behavior (\textit{i.e.}, motion) of drivers can be reasoned from their high level possible goals (\textit{i.e.}, intention) on the road.  To succeed in such behavior reasoning, we build a conditional prediction model to forecast goal-oriented trajectories with the following stages: (i) \textit{relational inference} where we encode relational interactions of vehicles using the perceptual context; (ii) \textit{intention estimation} to compute the probability distributions of intentional goals based on the inferred relations; and (iii) \textit{behavior reasoning} where we reason about the behaviors of vehicles as trajectories conditioned on the intentions. %To properly evaluate the performance of our approach, a new dataset is collected at road intersections. %with diverse interactions of vehicles.
    To this end, we extend the proposed framework to the pedestrian trajectory prediction task, showing the potential applicability toward general trajectory prediction.
\end{abstract}


%===============================================================================

\section{Introduction}
\label{sec:intro}

Forecasting  participants’ trajectories has gained huge attention in recent years. Extensive  research has focused on developing robotic systems for safe navigation in indoor and outdoor environments. In understanding human interaction, studies in \cite{alahi2016social,gupta2018social,Hasan_2018_CVPR,vemula2018social} have advanced our knowledge on pedestrian movement and social behaviors in crowded environments. Recent breakthroughs in automated driving technologies call for such research in the transportation domain. However, current knowledge on pedestrian behavior cannot be directly applied to predicting vehicle trajectories for the following reasons: (i) current models are based on human movement and thus may not directly be applicable to vehicles with faster speed; and (ii) road layouts, which can provide informative motion cues particularly in driving scenes, have been rarely considered in the literature. There have been some research efforts in the transportation community. However, they largely focus on highway scenarios~\cite{deo2018multi,park2018sequence}, relative trajectory of vehicles respective to ego-motion~\cite{yao2018egocentric,malla2019nemo,malla2020titan}, and ego-motion prediction~\cite{huang2019uncertainty}. Therefore, a robust solution still does not exist for driving environments.

Inspired by a study in~\cite{feinfield1999young} on the causation between intentions (\textit{i.e.}, cause) and behaviors (\textit{i.e.}, effect) of humans, we propose a trajectory prediction model using a relationship between future destinations (intentions) and intermediate locations (behaviors) of drivers. In real driving situations, humans are capable of estimating the intention of others based on prior interactions, which corresponds to the potential destination in future. Then, we subsequently anticipate intermediate paths with respect to the intention. From this viewpoint, automated driving or advanced driving assistance systems should be able to address the following questions: (i) Can they learn to estimate intentions and react to interactions with other vehicles using sensory data?; (ii) If so, how can the systems predict accurate trajectories under conditions of uncertain knowledge in a physically plausible manner?

\begin{figure}[t]
\begin{center}
 \includegraphics[width=1\textwidth]{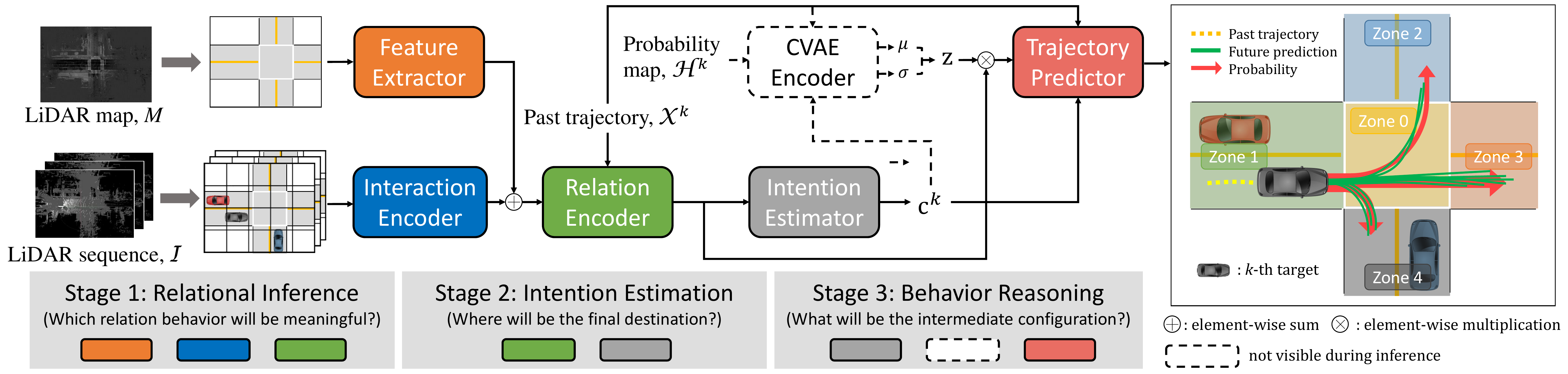}
\end{center}\vspace{-0.3cm}
   \caption{
    The proposed framework consists of 3 steps: 1) Infer the relational interactions of vehicles; 2) Estimate the probability distributions of intentional goals; 3) Conditionally reason about the goal-oriented behavior as trajectories being sampled from the estimated distributions. %The dotted lines are invisible during inference.
    %Predict the multi-modal trajectories of the vehicles conditioned on the intentions. 
    %The proposed approach forecasts future trajectories of vehicles. % by reasoning about their intention from relational behavior toward a surrounding environment. 
    %We first infer the relational interactions of vehicles with each other and with a given environment. The following module estimates the probability distribution of intentional goals (zones). Then, we conditionally reason about the goal-oriented behavior as trajectories being sampled from the estimated distribution. 
    }
\label{fig:main}\vspace{-0.6cm}
\end{figure}

Our framework, DROGON, is designed to address these questions. We infer relational interactions of vehicles with each other and with an environment. Based on this inference, we build a conditional probabilistic prediction model to forecast vehicle's goal-oriented trajectories. That is, we first estimate the probability distributions of intentions (\textit{i.e.}, potential destinations of vehicles as zones). Conditioned on the probabilities of formerly estimated intentions (\textit{e.g.}, 5 zones at a four-way intersection), we then predict the multi-modal trajectories of vehicles as illustrated in Fig.~\ref{fig:main}.

It is not feasible to demonstrate DROGON using the existing vehicle trajectory datasets ~\cite{colyar2007us101, colyar2007usi80, Geiger2012CVPR, ma2019trafficpredict, chang2019argoverse} since they do not provide the intentional destinations of individual agents. %Even if we were to add intention labels to these datasets, we would still need to understand the road topology of a scene at each time step with respect to the motion direction and orientation of all agents. Thus, adding zone labels to the existing datasets is impractical. 
Even adding intention labels to these datasets is not a
trivial task since we should understand road topology of the scene at each time step with respect to the motion direction and orientation of all agents. Thus, the zone label cannot be automatically added to any existing datasets.
Moreover, these datasets are restricted in their capacity to discover functional interactions between vehicles, in terms of the dataset size, motion diversity, and duration.
Therefore, we created Honda Intersection Dataset (HID)\footnote{\url{https://usa.honda-ri.com/hid}}, a large-scale vehicle trajectory prediction dataset to investigate the goal-oriented behavior, containing highly interactive scenarios at four-way intersections in the San Francisco Bay Area.% We extensively tested the proposed framework with the self-generated baselines as well as state-of-the-art methods using this dataset.

Furthermore, we extend DROGON to predict pedestrians' motion and demonstrate the applicability of our framework for general-purpose trajectory prediction. Unlike vehicles, the interactive environment of pedestrians would be hypothesized as a nearly open space as they move on paved / unpaved roads, sidewalks, grasslands, etc. We thus relax the assumption of intentional destinations to include various types of regions. We evaluate the extended model using the pedestrian trajectory benchmark datasets ~\cite{robicquet2016learning,pellegrini2009you,lerner2007crowds} to demonstrate the generalizability of the DROGON framework.

%The main contributions are summarized as follows: (i) Propose a trajectory forecast framework to estimate the intention of vehicles by analyzing their relational behavior; (ii) Reason causality between the intentional destination of an agent and its intermediate configuration for more accurate prediction; (iii) Create a new vehicle trajectory dataset with highly interactive scenarios at road intersections in urban areas and residential areas; and (iv) Present a way to generalize the proposed framework for pedestrian trajectory forecast. 
    
% The main contributions are summarized as follows: 
% \begin{enumerate}
%     \vspace{-0.2cm}
%     \item Propose a vehicle trajectory forecast framework to estimate the intention of vehicles by analyzing their relational behavior. 
%     \vspace{-0.2cm}
%     \item Reason causality between the intentional destination of an agent and its intermediate configuration for more accurate prediction.
%     \vspace{-0.2cm}
%     \item Create a new dataset at intersections with highly interactive scenarios in urban areas and residential areas.
%     \vspace{-0.2cm}
%     \item Present a way to generalize the framework for pedestrian trajectory forecast. 
%     \vspace{-0.2cm}
% \end{enumerate}

%-------------------------------------------------------------------------
\section{Related Work}

%We review the most relevant works on future trajectory prediction in the literature. We refer to \cite{ridel2018literature,rudenko2019human} for more general review on human behavior prediction.

\noindent
\textbf{Social interaction modeling} Following the pioneering work~\cite{helbing1995social,yamaguchi2011you}, there has been an explosion of research that has applied social interaction models to data-driven systems. Such models are basically trained using recurrent neural networks to make use of sequential attributes of human movements. In~\cite{alahi2016social}, a social pooling layer is introduced to model interactions of neighboring individuals, and \cite{gupta2018social, Hasan_2018_CVPR, yao2018egocentric, malla2019nemo,mangalam2020not,dwivedi2020ssp} improves its performance by using more efficient structure or adding supplemental cues. Recently, the recurrent operation is directly applied to interaction modeling. In~\cite{vemula2018social}, the relative importance of each person is captured using the attention mechanism, considering interactions between all humans. It is extended in~\cite{ma2019trafficpredict} with an assumption that the same types of road users show similar motion patterns. Although their predictions are acceptable in many cases, these approaches may fail in complex scenes without the perceptual consideration of the surrounding environment such as road structures or layouts.

\noindent
\textbf{Scene context as an additional modality} Scene context of an interacting environment has been presented in~\cite{lee2017desire} in addition to their social model. However, their restriction of the interaction boundary to local surroundings often causes failures toward far future prediction. \cite{xue2018ss,tang2019multiple,salzmann2020trajectron++,zhao2020tnt} subsequently extends local scene context through additional global scale image features. Also, \cite{choi2019looking} analyzes local scene context from a global perspective and encodes relational behavior of all agents. Motivated by their relational inference from the perceptual observation, we design a novel framework on top of relation-level behavior understanding. %Our approach takes advantage of relational inference for reasoning about the causal relationship between driver's intention and behavior.  

\noindent
\textbf{Goal-oriented trajectory prediction} The future motion of the target agent has been conditioned on the \textit{external} hypotheses such as the possible action of the neighboring agent in~\cite{rhinehart2019precog} or the potential motion of the ego-agent with respect to others in~\cite{malla2019nemo}. Although the proposed approach shares the similar aims for conditioning the trajectory, we directly explore the \textit{internal} intention of the target. Recently, a predefined trajectory set is used to represent the driver intention in~\cite{chai2019multipath}, which is restricted to generate trajectories outside of the anchor set. In contrast, we generate situation-aware behaviors (i.e., motion configuration) with help of the proposed generative pipeline conditioned on the driver's intentional goal (i.e., future destination).%construct a causation between intention (i.e., future destination) and behavior (i.e., motion configuration) of road agents for trajectory prediction.
%the potential path of the target agent in~\cite{chai2019multipath}. Particularly in the latter, they initially estimate a certain number of motion types from the predefined trajectory set (\textit{i.e.}, anchors), and then refine the initial estimates. As a result, their accuracy is heavily reliant on the quality of anchor generation. However, the proposed approach do not use such unreliable initialization, 

\noindent
\textbf{Trajectory datasets in driving scenes} The NGSIM~\cite{colyar2007us101,colyar2007usi80} dataset has been widely used in the transportation domain~\cite{li2019coordination,li2019grip} for vehicle trajectory forecast with different congestion levels in highways. However, the motion of vehicles and their interactions are mostly simple. The KITTI~\cite{Geiger2012CVPR} dataset includes multi-modal sensor data such as LiDAR point clouds, RGB images, and IMUs with various agent categories. However, its small number of tracklets makes the dataset barely used~\cite{lee2017desire} for the purpose of trajectory forecast. Recently, \cite{ma2019trafficpredict} released the dataset for the trajectory prediction task. It has been collected from urban driving scenarios as a subset of ApolloScape~\cite{huang2018apolloscape}. However, they only provide trajectory information with no corresponding visual data of ApolloScape, which is insufficient to discover visual scene context. Subsequently, the Argoverse~\cite{chang2019argoverse} forecasting dataset is available together with visual information such as 3D Maps and LiDAR data. Although its total number of motions is larger than KITTI, each segment is 5 $sec$ long, which results in the short-term (2 $sec$ observation and 3 $sec$ prediction) prediction horizon. For the general-purpose driving tasks, nuScenes~\cite{nuscenes2019} and Waymo~\cite{sun2019scalability} is recently introduced. Since they are not designed for trajectory prediction, the motion complexity is not considered. Therefore, we create a new trajectory dataset with more diverse vehicle motions in highly interactive scenarios, particularly at intersections.

\section{Preliminaries}

\subsection{Spatio-Temporal Interactions}

\begin{figure}[t]
\begin{center}
 \includegraphics[width=1\textwidth]{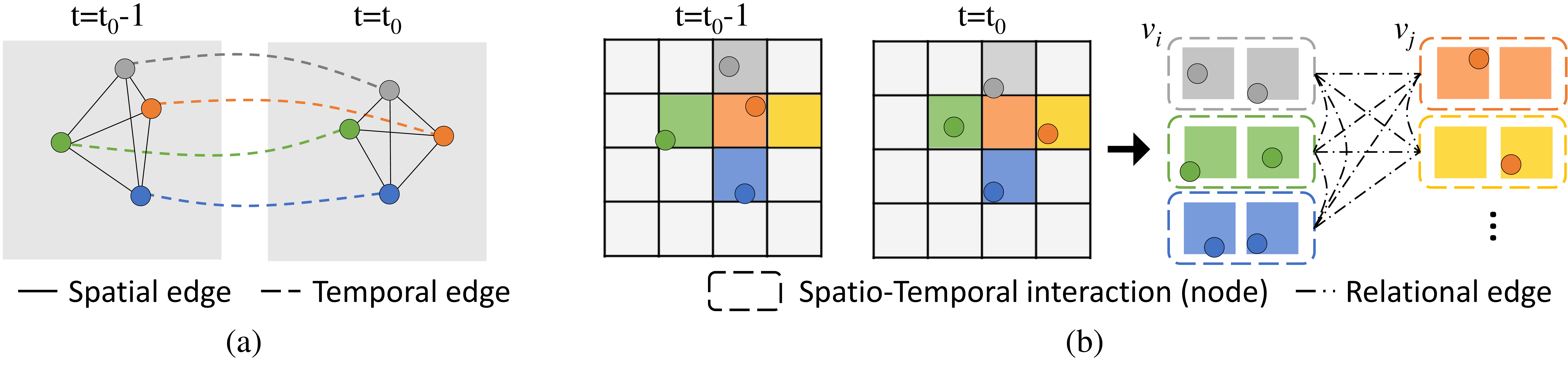}
\end{center}\vspace{-0.5cm}
   \caption{
   Illustration of different types of graph models to encode spatio-temporal interactions. (a) A node represents the state of each road user, whereas (b) it is a visual encoding of spatio-temporal interactions captured from each region of the discretized grid between adjacent frames.
   }
\label{fig:stgraph}\vspace{-0.3cm}
\end{figure}

Spatio-temporal interactions between road users have been considered as one of the most important features to understand their social behaviors. In~\cite{vemula2018social,ma2019trafficpredict}, spatio-temporal graph models are introduced with nodes to represent road users and edges to express their interactions with each other. To model spatio-temporal interactions, the spatial edges capture the relative motion of two nodes at each time step, and temporal edges capture the temporal motion of each node between adjacent frames as shown in Fig.~\ref{fig:stgraph}(a). Recently in~\cite{choi2019looking}, spatio-temporal features are visually computed using a convolutional kernel within a receptive field. In the spatio-temporal domain, these features not only contain interactions of road users with each other, but also incorporate their interactions with the environment. We use a similar approach and reformulate the problem with a graph model.  

\subsection{Relational Graph}

In the proposed approach, the traditional definition of a node is extended from an individual road user to a spatio-temporal feature representation obtained by exploiting spatial locality in input images. Thus, the edge captures relational behavior from spatio-temporal interactions of road users. We refer to this edge as `relational edge' as shown in Fig.~\ref{fig:stgraph}(b). In this view, we define an undirected and fully connected graph $\mathcal{G}=(\mathcal{V}, \mathcal{E})$, where $\mathcal{V}$ is a finite set of $|\mathcal{V}|=n$ nodes ($n=25$ is used) and $\mathcal{E}$ is a set of relational edges connecting each pair of nodes. Given $\tau$ number of input images, we visually extract a node $v_i\in \mathcal{V}$, where $v_i$ is a $d$-dimensional vector representing spatio-temporal interactions within the $i$-th region of the discretized grid. The feature $r_{ij}$ of the relational edge between two nodes $(v_i,v_j)$ first determines whether the given interaction pair has meaningful relations from a spatio-temporal perspective through the function $\phi$, and then the function $\theta$ is used to identify how their relations $r_{ij}$ can affect the future motion of the target $k$ based on its past motion context $q^k$: ${r_{ij}} = \phi(v_{ij};W^r)$ and ${f_{ij}^k} = \theta(r_{ij},q^k;W^f)$,
% \begin{align}
%   &\begin{aligned}
%     \mathllap{r_{ij}} &= \phi(v_{ij};W^r)\\
%   \end{aligned}\\
%   &\begin{aligned}
%     \mathllap{f_{ij}^k} &= \theta(r_{ij},q^k;W^f),\\
%   \end{aligned}
% \end{align}
where $v_{ij}= v_i\boxtimes v_j$ is the concatenation of two nodes, $W^r$ denotes the weight parameters of $\phi$, $W^f$ is those of $\theta$, and $q^k$ is an $m$-dimensional feature representation extracted from the past trajectory $\mathcal{X}^k = \{{X}^k_{t_0-\tau+1}, {X}^k_{t_0-\tau+2}, ..., {X}^k_{t_0}\}$ of the $k$-th agent observed in the given perceptual information. We subsequently collect relational information $f_{ij}^k$ from all pairs and perform element-wise sum to produce a unique relational representation $\mathcal{F}^k = \sum_{i,j}f_{ij}^k$ for the $k$-th agent.

%-------------------------------------------------------------------------
\section{Methodology}
We transfer knowledge of spatio-temporal relational inference $\mathcal{F}^k$ to predict the probability of intentional goals as well as goal-oriented trajectories. To accomplish this, we assemble building blocks from (i) \textit{relational inference} to encode relational interactions of vehicles using a relational graph, (ii) \textit{intention estimation} to compute the probability distribution of intentional goals based on the inferred relations from the perceptual context, and (iii) \textit{behavior reasoning} to reason about the goal-oriented behavior of drivers as future locations conditioned on the intentional destinations.

\subsection{Problem Definition}

%We assume raw LiDAR point clouds and 3D bounding boxes are already processed and projected into a top-down image coordinate. Details of preprocessing of raw data is described in Sec.~\ref{sec:preprocessing}. 
Given X$^k=\{\mathcal{I}, M, \mathcal{X}^k\}$, the proposed framework aims to predict $\delta$ number of likelihood heatmaps ${\mathcal{H}^k} = \{{H}_{t_0+1}^k,{H}_{t_0+2}^k, ..., {H}_{t_0+\delta}^k\}$ for the $k$-$th$ target vehicle observed in $\mathcal{I}$, where $\mathcal{I} = \{I_{t_0-\tau+1}, I_{t_0-\tau+2}, ..., I_{t_0}\}$ is $\tau$ number of past LiDAR images and $M$ is a top-down LiDAR map with a same coordinate with $\mathcal{I}$. Details are provided in the supplementary material. The future locations ${\mathcal{Y}^k} = \{{Y}_{t_0+1}^k,{Y}_{t_0+2}^k, ..., {Y}_{t_0+\delta}^k\}$ are found using a coordinate of a point with a maximum likelihood from each heatmap $H_t^k$.

\subsection{Behavior Reasoning for Trajectory Prediction}

\subsubsection{Conditional Trajectory Prediction}
We use a conditional VAE (CVAE) framework to forecast multiple possible trajectories of each vehicle. For given observation $c$, a latent variable $z$ is sampled from the prior distribution $P(z\vert c)$, and the output heatmaps $\mathcal{H}$ are generated from the distribution $P(\mathcal{H}\vert z, c)$. As a result, multiple $z$ drawn from the conditional distribution allows the system to model multiple outputs using the same observation $c$, where $c = q\boxtimes g$ is the concatenation of past motion context $q$ encoded from $\mathcal{X}$ and estimated intention $g$. In general, the true posterior $P(z\vert \mathcal{H},c)$ in maximum likelihood inference is intractable. Therefore, we consider an approximate posterior $Q(z\vert \mathcal{H},c)$ with variational parameters predicted by a neural network. %The variational lower bound of the model is thus written as follows: 
% \begin{equation}
%   \log P(\mathcal{H}\vert c) \geq - KL(Q(z\vert \mathcal{H},c)\Vert P(z\vert c)) + \mathbb{E}_{Q(z\vert \mathcal{H},c)}[\log P(\mathcal{H}|z,c)]
% \end{equation}
The objective of the model is thus written as follows:
\begin{equation}
  \mathcal{L}_{C} = - KL\left(Q(z\vert \mathcal{H},c)\Vert P(z\vert c)\right) + \mathbb{E}_{Q(z\vert \mathcal{H},c)}[\log P(\mathcal{H}|z,c)],
\end{equation}
where $z_l \thicksim Q(z_l\vert \mathcal{H},c) = \mathcal{N}%(\mu_l,\sigma_l)$ is modeled as Gaussian distribution with a mean $\mu_l$ and standard deviation $\sigma_l$.
(0,\textnormal{I})$ is modeled as Gaussian distribution.

We respectively build $Q(z\vert \mathcal{H},c)$ and $ P(\mathcal{H}|z,c)$ as a CVAE encoder and trajectory predictor, on top of convolutional neural networks. At training time, the observed condition $c$ is first concatenated with heatmaps $\mathcal{H}$, and we train the CVAE encoder to learn to approximate the prior distribution $P(z\vert c)$ by minimizing the Kullback-Leibler divergence. Once the model parameters are learned, the latent variable $z$ can be drawn from the same Gaussian distribution. At test time, the random sample $z\thicksim \mathcal{N}(0,\textnormal{I})$ is generated and masked with the relational features $\mathcal{F}$ using the element-wise multiplication operator. The resulting variable is passed through the trajectory predictor and concatenated with the observation $c$ to generate $\delta$ number of heatmaps $\widehat{\mathcal{H}}$. Details of the network architecture are described in the supplementary material.

\subsubsection{Intentional Goal Estimation}
\label{sec:intention}
We also train the intention estimator for goal-oriented future prediction which employs prior knowledge about the intention of vehicles (at time $t=t_0+\delta$). Given the relational features $\mathcal{F}$ extracted from vehicle interactions, we estimate the softmax probability $S_g$ for each intention category $g\in\{1,...,G\}$ ($G=5$ in Fig.~\ref{fig:main}) through a set of fully connected layers with a following ReLU activation function. We compute the cross-entropy from the softmax probability:
\begin{equation}
    \mathcal{L}_S = -\sum_{m=1}^{G} \mathbbm{1}(m=g) \log S_g,
\end{equation}
where $g$ is an estimated intention category and $\mathbbm{1}$ is the indicator function, which equals 1 if $m$ equals $g$ or 0 otherwise. We use the estimated intention $g$ to condition the process of model prediction. The computed softmax probability $S_g$ is later used at test time to sample $z$ with respect to its distribution.

\subsection{Explicit Penalty Modeling}

We introduce additional penalty terms specifically designed to constrain the model toward reliance on perceptual scene context and spatio-temporal priors.

\subsubsection{Penetration penalty} We encourage the model to forecast all future locations within a boundary of the drivable road in a given environment. To ensure that the predictions do not penetrate outside the road (\textit{i.e.}, sidewalks or buildings), we penalize the predicted points outside the drivable road using the following term:
\begin{equation}
  \mathcal{L}_P= \frac{1}{\delta}\sum_{t=t_0+1}^{t_0+\delta}\sum_{j=1}^{J}\left( \mathcal{D}_j\times B(\widehat{\mathcal{H}}_{t,j})\right), 
\end{equation}
where the function $B$ is the binary transformation with a threshold $\epsilon_B$, $\mathcal{D}$ is the binary mask annotated as zero inside the drivable road, and $J\in\mathbb{R}^{H\times W}$ is total pixels in each likelihood heatmap.
  
\subsubsection{Inconsistency penalty} In order to restrict our model from taking unrealistic velocity changes between adjacent frames, we encourage temporal consistency between frames as a way to smooth the predicted trajectories. We hypothesize that the current velocity at $t=t_0$ should be near to the velocity of both the previous frame ($t=t_0$-$1$) and next frame ($t=t_0$+$1$). The inconsistency penalty is defined as
\begin{equation}
  \mathcal{L}_I= \frac{1}{\delta-1}\sum_{t=t_0+1}^{t_0+\delta-1} E(\textnormal{v}_{t-1},\textnormal{v}_t,\textnormal{v}_{t+1}), 
\end{equation}
where v$_t$ denotes velocity at time $t$ and 
\begin{equation}
E(a,x,b) = \textnormal{max}(0,\textnormal{min}(a,b)-x) + \textnormal{max}(x-\textnormal{max}(a,b),0)
\end{equation}
is the term to softly penalize the predictions outside of the velocity range.

\subsubsection{Dispersion penalty} We further constrain the model to output more natural future trajectories, penalizing the cases where large prediction error is observed. In order to discourage the dispersion of an actual distance error distribution of the model, we use the following penalty:
\begin{equation}
  \mathcal{L}_D = \textnormal{Var}\left(\left\{\Vert{Y}_t - \widehat{{Y}}_t\Vert^2_2 \right\}_{t=t_0+1}^{t_0+\delta}\right) = \frac{1}{\delta}\sum_{t=t_0+1}^{t_0+\delta} (d_t - \bar{{d}})^2,
\end{equation}
where $d_t$ is an Euclidean distance between the predicted location and ground truth at time $t$ and $\bar{{d}}$ denotes a mean of $\boldsymbol{d} = \{d_{t_0+1},...,d_{t_0+\delta}\}$. We observe that the $\mathcal{L}_D$ penalty is particularly helpful to obtain accurate future locations with the concurrent use of the $\mathcal{L}_P$ term.

\subsection{Training}
At training time, we minimize the total loss drawn in Eqn.~\ref{eqn:10}. The first two terms are primarily used to optimize the CVAE modules which aims to approximate the prior and generate actual likelihood predictions. The third term mainly leads the model's output to be in the drivable road, and the last two terms are involved in generation of more realistic future locations. We set the loss weights as $\zeta = 1$, $\eta = 0.1$, and $\mu=0.01$ which properly optimized the entire network structures.
\begin{equation}
    \mathcal{L}_{Optimize} = -\mathcal{L}_C + \mathcal{L}_S + \zeta\mathcal{L}_P + \eta\mathcal{L}_I + \mu\mathcal{L}_D.
    \label{eqn:10}
\end{equation}

\subsection{Extension of DROGON}
\label{sec:ext}
At road intersections, we can define each potential destination in the scene as one of the zones based on its structural topology. In this way, each zone corresponds to the intentional destination of the driver as shown in Fig.~\ref{fig:main}. However, such a strategy cannot be directly applicable to pedestrians as their interactive environment is hypothesized as a nearly open space. The structural layout of the scene is not as informative as that of vehicles'. We thus relax the assumption for intentional destinations, so they can be any regions in the given environment for pedestrian trajectory prediction. By assuming every grid region in an image as zones, we can generalize the proposed behavior reasoning framework for pedestrian trajectory forecast in the open space. In the rest of this paper, we use a different abbreviation, DROGON-E, for the extended framework. Note that applying an extended method to driving scenes may cause a prediction failure since the future vehicle motion can be generated throughout non-drivable areas like sidewalks or buildings, as validated in Table~\ref{tbl:quan-m}.

%-------------------------------------------------------------------------
% \input{dataset}

%-------------------------------------------------------------------------

\section{Experiments}
We comprehensively evaluate the proposed approach using Honda Intersection Dataset (HID). The detailed specifications of HID can be found in the supplementary material. 

Although the authors are aware of other vehicle trajectory datasets such as %like ETH~\cite{pellegrini2009you}, UCY~\cite{lerner2007crowds}, and SDD~\cite{robicquet2016learning} for pedestrians or 
~\cite{Geiger2012CVPR,ma2019trafficpredict,chang2019argoverse}, we do not use them for one or more of the following reasons: (i) None of the datasets provides the structure-specific intentional destinations of agents. Such zone labels should be acquired by hand-labeling as explained in Sec.~\ref{sec:intro}. It thus makes the demonstration of DROGON infeasible to reason about behaviors conditioned on \textit{internal} intentions; and (ii) Perceptual information such as RGB images or LiDAR point clouds is not provided, which is critical to visually infer relational behavior between agents from our framework. %; (iii) The dataset is designed for general-purpose autonomous driving tasks, which results in rather simple vehicle motions for the trajectory prediction problem. 
We did not find a straightforward way to evaluate our behavior reasoning framework on these datasets. 

Additionally, we evaluate the extended framework DROGON-E for 
%Then, we provide a solution to extend the use of DROGON for 
pedestrian trajectory prediction. Its generalization is validated using three public datasets (SDD~\cite{robicquet2016learning}, ETH~\cite{pellegrini2009you}, and UCY~\cite{lerner2007crowds}) that contain pedestrian trajectories in diverse interaction scenarios. %Note that applying such an extended method to vehicle trajectory prediction may results in the failure of prediction since the future vehicle motion can be generated throughout any non-drivable regions like sidewalks or buildings.

\subsection{Comparison to Baselines}
\label{sec:baselines}

We conduct ablative tests using HID dataset to demonstrate the efficacy of the proposed DROGON framework by measuring average distance error (ADE) during a given time interval and final distance error (FDE) at a specific time frame in \textit{meters}. %For this, we compare DROGON to the self-generated baselines by conducting evaluation using our intersection dataset. 

\begin{table}[t]%[!htb]
    % \caption{Global caption}
    \begin{minipage}{.48\linewidth}
      \caption{Quantitative comparison (ADE / FDE in \textit{meters}) for \textbf{single-modal} prediction. 
   }
      \centering
        \resizebox{1 \textwidth}{!}{   %resize table
        \begin{tabular}{l|lllll}
\hline
Single-modal&1.0 $sec$&2.0 $sec$&3.0 $sec$&4.0 $sec$\\
\hline 
\hline
% \topmidheader{1}{State-of-the-art methods}
\textit{State-of-the-art}\\
~{S-LSTM}\cite{alahi2016social} &1.66 / 2.18 &2.57 / 4.03&3.59 / 6.19&4.61 / 8.45 \\%,done
~{S-GAN}\cite{gupta2018social} &1.61 / 3.01 &2.06 / 3.83&2.32 / 4.35&4.28 / 7.92 \\%,done
~{S-ATTN}\cite{vemula2018social}&1.17 / 1.45 &1.69 / 2.61&2.41 / 4.45&3.29 / 6.67 \\%,done
~{Const-Vel}\cite{scholler2019simpler} &0.52 / 0.85 &1.27 / 2.63&2.34 / 5.38&3.70 / 8.88 \\%, done
~{Gated-RN}\cite{choi2019looking} &0.74 / 0.98 &1.14 / 1.79&1.60 / 2.89&2.13 / 4.20 \\%800,done
\hline
% \topmidheader{1}{Ours}
\textit{Ours}\\
~{DROGON}&\textbf{0.52} / \textbf{0.71} &\textbf{0.86} / \textbf{1.46}&\textbf{1.31} / \textbf{2.60}&\textbf{1.86} / \textbf{4.02} \\ %810_1,done
\hline
% \topmidheader{5}{Self-generated baselines}
\textit{Baseline}\\
~{w/o Intention}&0.79 / 1.04 &1.20 / 1.85&1.65 / 2.90&2.18 / 4.25 \\%802,done
~{w/o Map}&0.65 / 0.86 &1.01 / 1.62&1.46 / 2.77&2.02 / 4.23 \\%808,done
~{w/o Penalty}&0.60 / 0.81 &0.97 / 1.58&1.41 / 2.71&1.98 / 4.20 \\%807,done
\hline
\end{tabular}}
\label{tbl:quan-s}%\vspace{-0.5cm}
    \end{minipage}%
    \quad
    \begin{minipage}{.5\linewidth}
    \vspace{-0.55cm}
      \centering
        \caption{Quantitative comparison (ADE / FDE in \textit{meters}) for \textbf{multi-modal} prediction. }
        \scriptsize
        \resizebox{1 \textwidth}{!}{
        \begin{tabular}{l|lllll}
\hline
Multi-modal&1.0 $sec$&2.0 $sec$&3.0 $sec$&4.0 $sec$\\
\hline 
\hline
\textit{State-of-the-art}\\
~{S-LSTM}~\cite{alahi2016social} &1.06 / 1.37 &1.68 / 2.79&2.46 / 4.55&3.36 / 6.73 \\%,done
~{S-GAN}~\cite{gupta2018social} &1.50 / 2.84 &1.94 / 3.52&1.99 / 3.75&3.43 / 6.47\\%,done
~{S-ATTN}~\cite{vemula2018social}&1.35 / 1.69 &1.73 / 2.10&2.09 / 3.11&2.66 / 5.10\\%,done
~{Gated-RN}~\cite{choi2019looking} &0.60 / 0.80 &0.93 / 1.49&1.33 / 2.48&1.82 / 3.74\\%800
\hline
\textit{Ours}\\
~{DROGON-Best}&0.39 / 0.53 &0.65 / 1.14&1.03 / 2.11&1.48 / 3.29\\%810_1,done
~{DROGON-Prob}&\textbf{0.38} / \textbf{0.49} &\textbf{0.55} / \textbf{0.84}&\textbf{0.77} / \textbf{1.40}&\textbf{1.05} / \textbf{2.25}\\
\hline
\textit{Baseline}\\
~{DROGON-E}& 0.41 / 0.77 &0.84 / 1.53 &1.33 / 2.57 & 1.87 / 3.38 \\
\hline
\end{tabular}
}
\label{tbl:quan-m}%\vspace{-0.2cm}
    \end{minipage} \vspace{-0.5cm}
\end{table}

\noindent
\textbf{Prior knowledge of intention} In order to investigate the efficacy of behavior reasoning, we design a baseline (w/o Intention) by dropping the intention estimator and CVAE encoder from DROGON. As a result, this baseline is not generative, outputting a single set of deterministic locations. In Table~\ref{tbl:quan-s}, the reported error rates indicate that behavior reasoning is essential to predict accurate trajectories under conditions of prior knowledge of intention. It is due to the fact that goal-oriented reasoning is practically helpful to condition the search space and guide the course of future motion. The mean average precision of intention estimation is 71.1\% (from DROGON) and 70.2\% (from w/o map).

\noindent
\textbf{Global scene context} We define another baseline model (w/o Map) which does not use global scene context for trajectory forecast. For implementation, we did not add features extracted from the map $M$ into the relational inference stage. In this way, the model is not guided to learn global road layouts, similar to relational inference in~\cite{choi2019looking}. As shown in Table~\ref{tbl:quan-s}, the prediction error of this baseline definitely increases against DROGON. The comparison indicates that discovering additional global context encourages the model to better understand about the spatial environment.

\begin{figure*}[t]
  \centering
  \begin{subfigure}[b]{0.2\linewidth}
    \includegraphics[width=\linewidth]{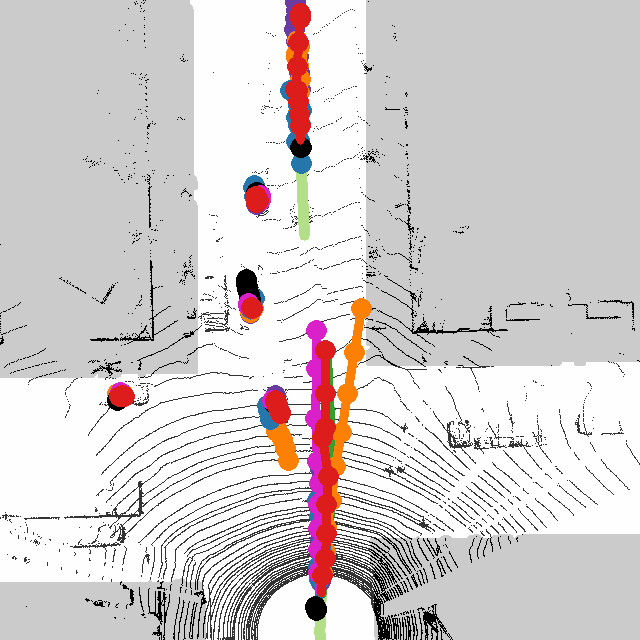}
     \caption{}\label{fig:3a}
  \end{subfigure}\quad
  \begin{subfigure}[b]{0.2\linewidth}
    \includegraphics[width=\linewidth]{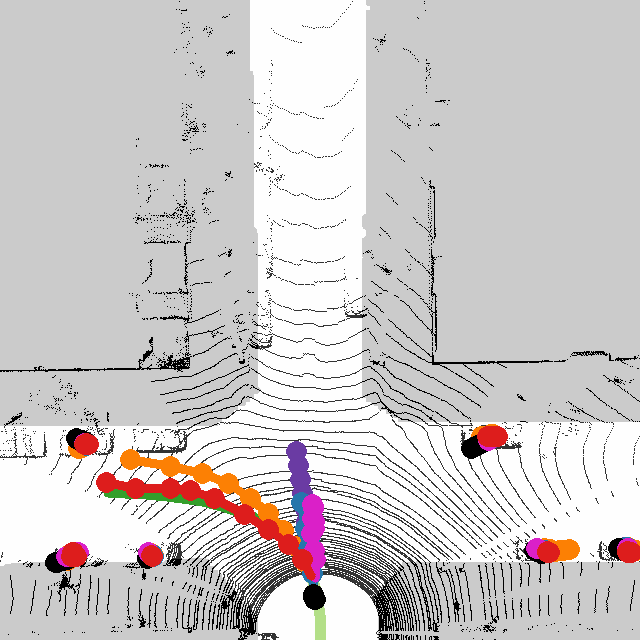}
    \caption{}\label{fig:3b}
  \end{subfigure}\quad
  \begin{subfigure}[b]{0.2\linewidth}
    \includegraphics[width=\linewidth]{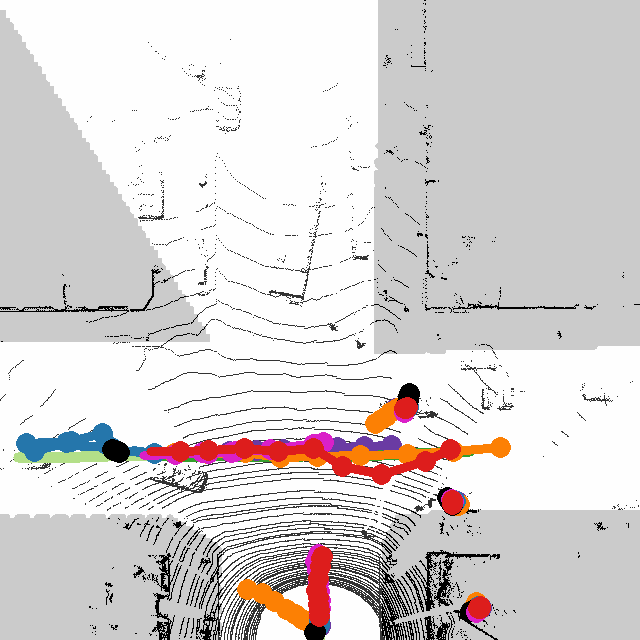}
    \caption{}\label{fig:3c}
  \end{subfigure}\quad
  \begin{subfigure}[b]{0.17\linewidth}
    \includegraphics[width=\linewidth]{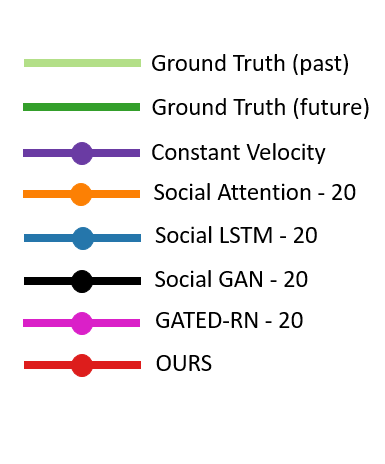}
  \end{subfigure}\vspace{-0.3cm}
  \caption{Qualitative comparison of DROGON with the state-of-the-art algorithms. We visualize the top-1 prediction. Gray mask is shown for non-drivable region.
  }
  \label{fig:others}\vspace{-0.2cm}
\end{figure*}
\begin{figure*}[!t]
  \centering
  \begin{subfigure}[b]{0.17\linewidth}
    \includegraphics[width=\linewidth]{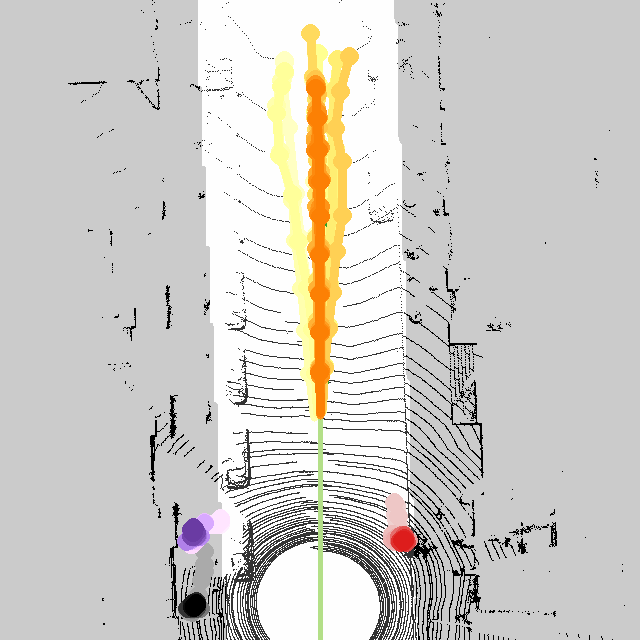}
     \caption{}\label{fig:4a}
  \end{subfigure}\quad
  \begin{subfigure}[b]{0.17\linewidth}
    \includegraphics[width=\linewidth]{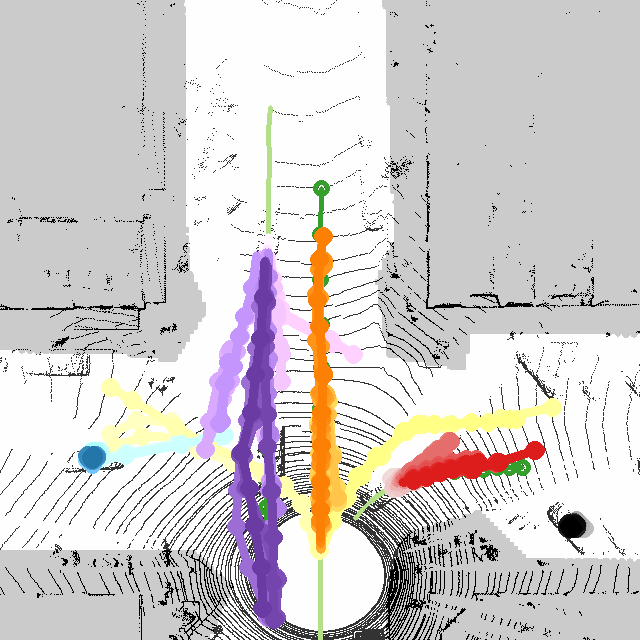}
    \caption{}\label{fig:4b}
  \end{subfigure}\quad
  \begin{subfigure}[b]{0.17\linewidth}
    \includegraphics[width=\linewidth]{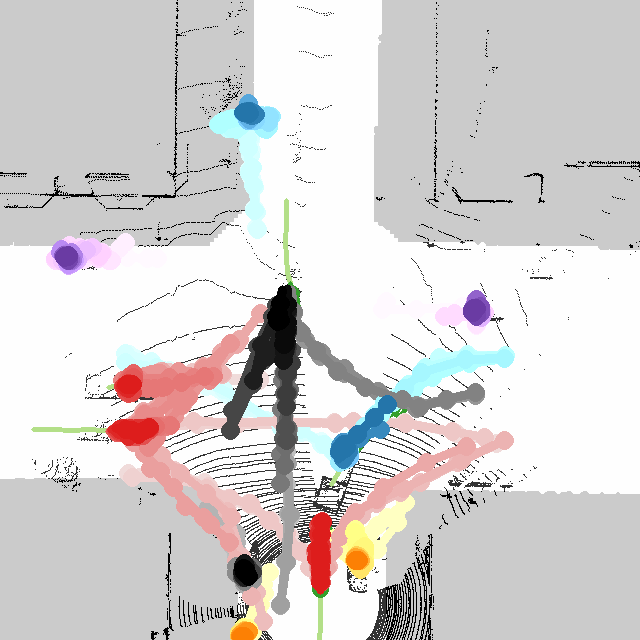}
    \caption{}\label{fig:4c}
  \end{subfigure}\quad
  \begin{subfigure}[b]{0.17\linewidth}
    \includegraphics[width=\linewidth]{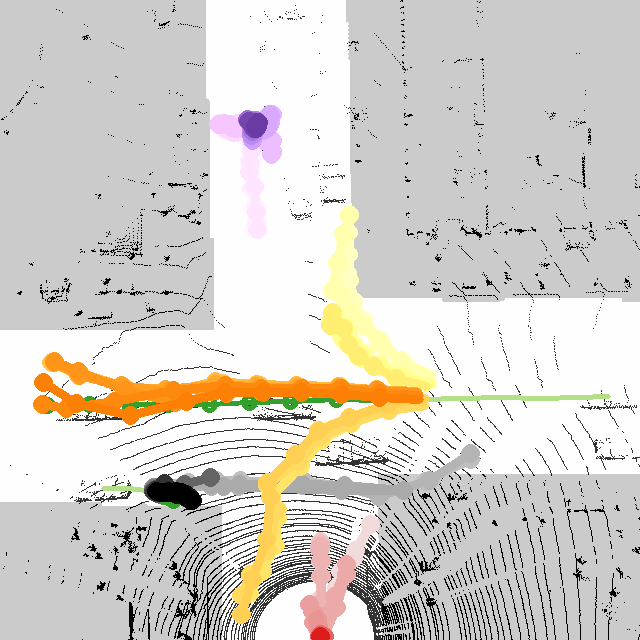}
    \caption{}\label{fig:4d}
  \end{subfigure}\quad
  \begin{subfigure}[b]{0.17\linewidth}
    \includegraphics[width=\linewidth]{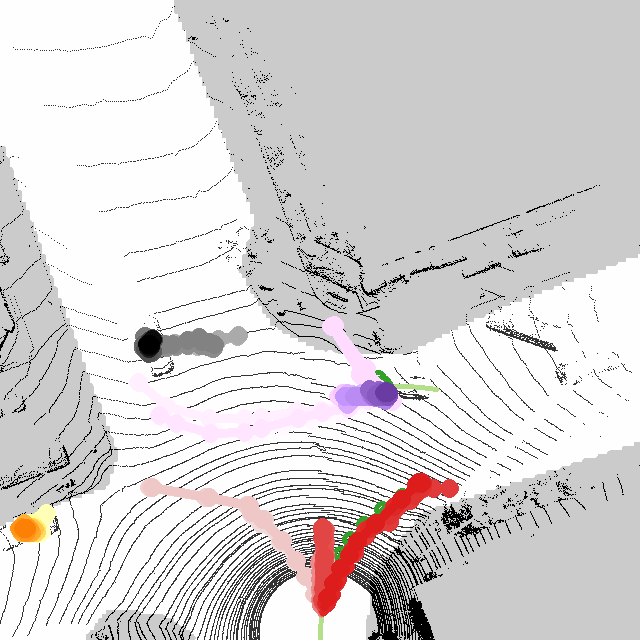}
     \caption{}\label{fig:4e}
  \end{subfigure}
%   \quad
%   \begin{subfigure}[b]{0.23\linewidth}
%     \includegraphics[width=\linewidth]{figures/201706061647_10760_11480_469_00.png}
%     \caption{}\label{fig:4f}
%   \end{subfigure}\quad
% \begin{subfigure}[b]{0.23\linewidth}
%     \includegraphics[width=\linewidth]{figures/201709151525_26380_26620_51_00.png}
%     \caption{}\label{fig:4g}
%   \end{subfigure}\quad
%   \begin{subfigure}[b]{0.23\linewidth}
%     \includegraphics[width=\linewidth]{figures/201709211444_28119_28350_93_00.png}
%     \caption{}\label{fig:4h}
%   \end{subfigure}
  \vspace{-0.2cm}
  \caption{(a-e) All 20 trajectories of DROGON-Prob-20 are plotted in interactive scenarios at interactions. We change the intensity of colors for those 20 samples and use different colors for different vehicles. Gray mask is shown for non-drivable region. }
  \label{fig:ours}\vspace{-0.6cm}
\end{figure*}

\noindent
\textbf{Explicit penalty} We now remove the penalty terms in the total loss from the proposed DROGON framework at training time. The performance of this baseline model (w/o Penalty) is compared in Table~\ref{tbl:quan-s}. Although its performance is higher than other baseline models, it achieves higher error rate in comparison to DROGON. % indicate that this baseline often generates unacceptable future trajectories. 
This is apparent in the sense that the model is not explicitly guided by physical constraints of the real world. Thus, we conclude that these penalty terms are dominant in forecasting accurate future trajectories.

\subsection{Comparison with the State of the Arts}
We compare the performance of DROGON to several state-of-the-art trajectory prediction approaches~\cite{alahi2016social,gupta2018social,vemula2018social,choi2019looking} that have shown outstanding performance for vehicle trajectory forecast~\cite{ma2019trafficpredict,messaoud2019non,roy2019vehicle,bi2019joint}. Extensive evaluations are conducted on tasks for both single-modal and multi-modal prediction. As shown in Table~\ref{tbl:quan-s} for single trajectory prediction, the performance of S-GAN~\cite{gupta2018social} is consistently improved against S-LSTM~\cite{alahi2016social} all over the time steps. S-ATTN~\cite{vemula2018social} shows further improvement of both ADE and FDE by employing relative importance of individual vehicles. Interestingly, however, their performance is worse than or comparable to the simple constant velocity (Const-Vel) model in~\cite{scholler2019simpler}. % without using any environmental priors. It is also shown true that the network model (Gated-RN in~\cite{choi2019looking}) can then better perform by adopting perceptual information about the interactive environment.
With additional perceptual priors, the network model (Gated-RN in~\cite{choi2019looking}) then performs better than the heuristic approach. DROGON also employs visual information of the physical environment. Additionally, we generate intentional goals and predict a trajectory by reasoning about goal-oriented behavior. As a result, we achieve the best performance against the state-of-the-art counterparts.

For evaluation on multi-modal prediction in Table~\ref{tbl:quan-m}, we generate $S=20$ samples and report an error of the $s$-$th$ prediction with minimum ADE (\textit{i.e. }, $\min_{s\in S}\Vert \mathcal{Y}^k- \widehat{\mathcal{Y}}_{s}^k\Vert^2_2$) as proposed in~\cite{lee2017desire,gupta2018social}. We design two variants of DROGON with a different sampling strategy: (i) DROGON-Best-20 generates trajectories only conditioned on the best intention estimate; and (ii) DROGON-Prob-20 conditions the model proportional to the softmax probability $S_g$ of each intention category. Similar to single-modal prediction, our models show a lower error rate than that of other approaches. It validates the effectiveness of our behavior reasoning framework for goal-oriented future forecast. In Fig.~\ref{fig:others}, we display their qualitative comparison in general driving scenarios \ref{fig:3a}, by considering the influence of environments (parked cars and road layouts) while making turns \ref{fig:3b}, and with an ability to socially avoid potential collisions \ref{fig:3c}. DROGON properly forecasts trajectories considering interactions with other vehicles and the environment. Moreover, we achieve the best performance with DROGON-Prob-20. By taking adaptive condition on potential goals, we can eventually ease the impact of misclassification in intention estimation. In Fig.~\ref{fig:ours}, we visualize goal-oriented trajectories reasoned from DROGON-Prob-20. While approaching \ref{fig:4a} and passing the intersection \ref{fig:4b}-\ref{fig:4e}, DROGON accordingly predicts goal-oriented trajectories using the intentional destination (zone) of vehicles. Note that our framework is able to predict future dynamic motion of the static vehicles (red and purple in \ref{fig:4d}), which can eventually help to avoid potential collisions that might be caused by their unexpected motion.

\begin{table*}[!t]
\footnotesize
\centering
 \caption{Quantitative comparison of DROGON-E with state-of-the-art methods using the SDD~\cite{robicquet2016learning} dataset. In a range of 1 - 4 sec, FDE is reported in \textit{pixels} at 1/5 resolution following \cite{lee2017desire,choi2019looking}. For 4.8 sec in the future, both ADE and FDE are reported using the original resolution as in \cite{sadeghian2018car,chai2019multipath}.}

    \begin{tabular}{cc||cccccc|c}
\hline
Time & Metric&CVAE & S-LSTM & DESIRE & CAR-Net & Gated-RN& MultiPath & Ours\\
\hline 
\hline
1.0 $sec$& \multirow{4}{*}{FDE}& 1.84 & 3.38 & 1.29 & - & 2.11 & - &\textbf{1.24} \\
2.0 $sec$& & 3.93 & 5.33 & 2.35 & - & 3.83 & - &\textbf{2.19}   \\
3.0 $sec$& & 6.47 & 9.58 & 3.47 & - & 5.98 & - &\textbf{3.36} \\
4.0 $sec$& &9.65 & 14.57& 5.33 & - & 8.65 & - &\textbf{4.94} \\
\hline
\hline
% 4.8 & 30.91 / 61.40& 31.19 / 56.97 & 19.25 / 34.05 &25.72 / 51.80 & 26.67 / 53.93 & 17.51 / 58.38 &\textbf{17.06 / 30.90} \\
\multirow{2}{*}{4.8 $sec$}& ADE & 30.91 & 31.19 & 19.25  &25.72  & 26.67  & 17.51 &~\textbf{17.06 } \\
& FDE&61.40& 56.97 & 34.05 &51.80 & 53.93 & 58.38 &\textbf{30.90} \\
\hline
\end{tabular}%\vspace{-0.2cm}
\label{tbl:quan-sdd}%\vspace{-0.4cm}
\end{table*}

\begin{table*}[!t]
\footnotesize
\centering
\captionof{table}{Quantitative comparison (ADE / FDE in \textit{meters}) of the proposed approach (DROGON-E) with the state-of-the-art methods -- S-GAN~\cite{gupta2018social}, SoPhie~\cite{sadeghian2018sophie}, S-BiGAT~\cite{kosaraju2019social}, PMP-NMMP~\cite{hu2020collaborative}, S-STGCNN~\cite{mohamed2020social} -- using the ETH~\cite{pellegrini2009you} and UCY~\cite{lerner2007crowds} dataset. }
    \begin{tabular}{l||lllll|l}
\hline
&ETH\_hotel&ETH\_eth&UCY\_univ&UCY\_zara01&UCY\_zara02&Average\\
\hline 
\hline
S-GAN~\cite{gupta2018social}& 0.87 / 1.62 &0.67 / 1.37&0.76 / 1.52&0.35 / 0.68&0.42 / 0.84&0.61 / 1.21\\
SoPhie~\cite{sadeghian2018sophie}& 0.70 / 1.43 &0.76 / 1.67&0.54 / 1.24&\textbf{0.30} / 0.63&0.38 / 0.78&0.54 / 1.15 \\
S-BiGAT~\cite{kosaraju2019social}& 0.69 / 1.29 & 0.49 / 1.01 & 0.55 / 1.32 & \textbf{0.30} / 0.62 & 0.36 / 0.75 & 0.48 / 1.00\\
PMP-NMMP~\cite{hu2020collaborative}& \textbf{0.61} / 1.08 & 0.33 / 0.63 & \textbf{0.52} / 1.11 & 0.32 / 0.66 & 0.29 / 0.61 & 0.41 / 0.82\\
S-STGCNN~\cite{mohamed2020social}& 0.64 / 1.11 & 0.49 / 0.85 & 0.44 / \textbf{0.79} & 0.34 / \textbf{0.53} & 0.30 / 0.48 & 0.44 / 0.75\\
\hline
{Ours}&0.68 / \textbf{0.95} &\textbf{0.10} / \textbf{0.16} &0.53 / 0.89 &0.45 / 0.81&\textbf{0.27} / \textbf{0.46}&\textbf{0.41} / \textbf{0.65} \\
%0.011 / 0.018 | 0.022 / 0.031 | 0.035 / 0.059 | 0.030 / 0.054 | 0.018 / 0.031 | 0.020 / 0.033
\hline
\end{tabular}%\vspace{-0.2cm}
\label{tbl:quan2}\vspace{-0.4cm}
\end{table*}

\subsection{Generalization of DROGON}
As detailed in Sec.~\ref{sec:ext}, we assume every grid region in an image as zones to generalize the proposed behavior reasoning framework. In this way, we further conduct cross-domain validation by evaluating DROGON-E on the widely used benchmark datasets for pedestrian trajectory forecast.  %The SDD dataset~\cite{robicquet2016learning} contains diverse motion of pedestrians, cyclists, skateboarders, carts, etc. in 60 videos, captured in the various locations of the university campus using the drone-mounted camera. 
We first use the SDD dataset to evaluate the proposed framework comparing with the current state-of-the-art methods~\cite{alahi2016social,lee2017desire,sadeghian2018car,choi2019looking,chai2019multipath} on two standard benchmark measures, (i) FDE at 1-4 $sec$ as used in~\cite{lee2017desire,choi2019looking} and (ii) ADE / FDE at 4.8 $sec$ as reported in~\cite{sadeghian2018car,chai2019multipath}. For evaluation, we divide the original image space into $5\times5$ regions ($G=25$), assuming the intentional destination of the target agent belongs to one of regions. As shown in Table~\ref{tbl:quan-sdd}, DROGON-E achieves the best performance over all time steps compared to the state-of-the-art methods. It validates the efficacy of the proposed behavior reasoning framework. 
%We next evaluate DROGON-E using the ETH and UCY datasets. The same number of intention categories ($G=25$) are assumed as goals. Following the evaluation metric in~\cite{xue2018ss,choi2019looking,dwivedi2020ssp}, we report ADE / FDE at 4.8 $sec$ in normalized $pixels$ in Table~\ref{tbl:quan2}. DROGON-E consistently outperforms the state-of-the-art methods~\cite{alahi2016social,xue2018ss,gupta2018social,choi2019looking,dwivedi2020ssp}. These results further validate the generalization capability of our approach toward pedestrian trajectory prediction.
We next evaluate DROGON-E using the ETH and UCY datasets. The same number of intention categories ($G=25$) are assumed as goals. We report ADE / FDE at 4.8 $sec$ in $meters$ in Table~\ref{tbl:quan2}. DROGON-E outperforms the state-of-the-art methods~\cite{gupta2018social,sadeghian2018sophie,kosaraju2019social,hu2020collaborative,mohamed2020social} from hotel, eth, and zara02 subset, improving average errors. These results further validate the generalization capability of our approach toward pedestrian trajectory prediction.

\section{Conclusion}
\label{sec:conclusion}

We presented a Deep RObust Goal-Oriented trajectory prediction Network, DROGON, which aims to predict a behavior of human drivers conditioned on their intentions. Motivated by the real world scenarios, the proposed framework estimates the intention of drivers based on their relational behavior. Given prior knowledge of intention, DROGON reasons about the behavior of vehicles as intermediate paths. To this end, multiple possible trajectories of each vehicle are generated considering physical constraints of the real world. For comprehensive evaluation, we collected a large-scale dataset with highly interactive scenarios at intersections and tested DROGON comparing with the current state-of-the-art methods. We further provided a way to generalize the proposed framework for pedestrian trajectory prediction, which also validates the efficacy of behavior reasoning. 

%===============================================================================

% The maximum paper length is 8 pages excluding references and acknowledgements, and 10 pages including references and acknowledgements

\clearpage
% The acknowledgments are automatically included only in the final version of the paper.
% \acknowledgments{If a paper is accepted, the final camera-ready version will (and probably should) include acknowledgments. All acknowledgments go at the end of the paper, including thanks to reviewers who gave useful comments, to colleagues who contributed to the ideas, and to funding agencies and corporate sponsors that provided financial support.}

%===============================================================================

\bibliography{example}  % .bib

\begin{thebibliography}{45}
\providecommand{\natexlab}[1]{#1}
\providecommand{\url}[1]{\texttt{#1}}
\expandafter\ifx\csname urlstyle\endcsname\relax
  \providecommand{\doi}[1]{doi: #1}\else
  \providecommand{\doi}{doi: \begingroup \urlstyle{rm}\Url}\fi

\bibitem[Alahi et~al.(2016)Alahi, Goel, Ramanathan, Robicquet, Fei-Fei, and
  Savarese]{alahi2016social}
A.~Alahi, K.~Goel, V.~Ramanathan, A.~Robicquet, L.~Fei-Fei, and S.~Savarese.
\newblock Social lstm: Human trajectory prediction in crowded spaces.
\newblock In \emph{CVPR}, 2016.

\bibitem[Gupta et~al.(2018)Gupta, Johnson, Fei-Fei, Savarese, and
  Alahi]{gupta2018social}
A.~Gupta, J.~Johnson, L.~Fei-Fei, S.~Savarese, and A.~Alahi.
\newblock Social gan: Socially acceptable trajectories with generative
  adversarial networks.
\newblock In \emph{CVPR}, 2018.

\bibitem[Hasan et~al.(2018)Hasan, Setti, Tsesmelis, Del~Bue, Galasso, and
  Cristani]{Hasan_2018_CVPR}
I.~Hasan, F.~Setti, T.~Tsesmelis, A.~Del~Bue, F.~Galasso, and M.~Cristani.
\newblock Mx-lstm: Mixing tracklets and vislets to jointly forecast
  trajectories and head poses.
\newblock In \emph{CVPR}, 2018.

\bibitem[Vemula et~al.(2018)Vemula, Muelling, and Oh]{vemula2018social}
A.~Vemula, K.~Muelling, and J.~Oh.
\newblock Social attention: Modeling attention in human crowds.
\newblock In \emph{ICRA}, 2018.

\bibitem[Deo and Trivedi(2018)]{deo2018multi}
N.~Deo and M.~M. Trivedi.
\newblock Multi-modal trajectory prediction of surrounding vehicles with
  maneuver based lstms.
\newblock In \emph{IV}, 2018.

\bibitem[Park et~al.(2018)Park, Kim, Kang, Chung, and Choi]{park2018sequence}
S.~H. Park, B.~Kim, C.~M. Kang, C.~C. Chung, and J.~W. Choi.
\newblock Sequence-to-sequence prediction of vehicle trajectory via lstm
  encoder-decoder architecture.
\newblock In \emph{IV}, 2018.

\bibitem[Yao et~al.(2019)Yao, Xu, Choi, Crandall, Atkins, and
  Dariush]{yao2018egocentric}
Y.~Yao, M.~Xu, C.~Choi, D.~J. Crandall, E.~M. Atkins, and B.~Dariush.
\newblock Egocentric vision-based future vehicle localization for intelligent
  driving assistance systems.
\newblock In \emph{ICRA}, 2019.

\bibitem[Malla et~al.(2020{\natexlab{a}})Malla, Dwivedi, Dariush, and
  Choi]{malla2019nemo}
S.~Malla, I.~Dwivedi, B.~Dariush, and C.~Choi.
\newblock Nemo: Future object localization using noisy ego priors.
\newblock \emph{arXiv preprint arXiv:1909.08150}, 2020{\natexlab{a}}.

\bibitem[Malla et~al.(2020{\natexlab{b}})Malla, Dariush, and
  Choi]{malla2020titan}
S.~Malla, B.~Dariush, and C.~Choi.
\newblock Titan: Future forecast using action priors.
\newblock In \emph{CVPR}, 2020{\natexlab{b}}.

\bibitem[Huang et~al.(2019)Huang, McGill, Williams, Fletcher, and
  Rosman]{huang2019uncertainty}
X.~Huang, S.~McGill, B.~C. Williams, L.~Fletcher, and G.~Rosman.
\newblock Uncertainty-aware driver trajectory prediction at urban
  intersections.
\newblock \emph{ICRA}, 2019.

\bibitem[Feinfield et~al.(1999)Feinfield, Lee, Flavell, Green, and
  Flavell]{feinfield1999young}
K.~A. Feinfield, P.~P. Lee, E.~R. Flavell, F.~L. Green, and J.~H. Flavell.
\newblock Young children's understanding of intention.
\newblock \emph{Cognitive Development}, 1999.

\bibitem[Colyar and Halkias(2007{\natexlab{a}})]{colyar2007us101}
J.~Colyar and J.~Halkias.
\newblock Us highway 101 dataset.
\newblock \emph{Federal Highway Administration (FHWA), Tech. Rep.
  FHWA-HRT-07-030}, 2007{\natexlab{a}}.

\bibitem[Colyar and Halkias(2007{\natexlab{b}})]{colyar2007usi80}
J.~Colyar and J.~Halkias.
\newblock Us highway i-80 dataset.
\newblock \emph{Federal Highway Administration (FHWA), Tech. Rep.
  FHWA-HRT-07-030}, 2007{\natexlab{b}}.

\bibitem[Geiger et~al.(2012)Geiger, Lenz, and Urtasun]{Geiger2012CVPR}
A.~Geiger, P.~Lenz, and R.~Urtasun.
\newblock Are we ready for autonomous driving? the kitti vision benchmark
  suite.
\newblock In \emph{CVPR}, 2012.

\bibitem[Ma et~al.(2019)Ma, Zhu, Zhang, Yang, Wang, and
  Manocha]{ma2019trafficpredict}
Y.~Ma, X.~Zhu, S.~Zhang, R.~Yang, W.~Wang, and D.~Manocha.
\newblock Trafficpredict: Trajectory prediction for heterogeneous
  traffic-agents.
\newblock In \emph{AAAI}, 2019.

\bibitem[Chang et~al.(2019)Chang, Lambert, Sangkloy, Singh, Bak, Hartnett,
  Wang, Carr, Lucey, Ramanan, et~al.]{chang2019argoverse}
M.-F. Chang, J.~Lambert, P.~Sangkloy, J.~Singh, S.~Bak, A.~Hartnett, D.~Wang,
  P.~Carr, S.~Lucey, D.~Ramanan, et~al.
\newblock Argoverse: 3d tracking and forecasting with rich maps.
\newblock In \emph{CVPR}, 2019.

\bibitem[Robicquet et~al.(2016)Robicquet, Sadeghian, Alahi, and
  Savarese]{robicquet2016learning}
A.~Robicquet, A.~Sadeghian, A.~Alahi, and S.~Savarese.
\newblock Learning social etiquette: Human trajectory understanding in crowded
  scenes.
\newblock In \emph{ECCV}, 2016.

\bibitem[Pellegrini et~al.(2009)Pellegrini, Ess, Schindler, and
  Van~Gool]{pellegrini2009you}
S.~Pellegrini, A.~Ess, K.~Schindler, and L.~Van~Gool.
\newblock You'll never walk alone: Modeling social behavior for multi-target
  tracking.
\newblock In \emph{ICCV}, 2009.

\bibitem[Lerner et~al.(2007)Lerner, Chrysanthou, and
  Lischinski]{lerner2007crowds}
A.~Lerner, Y.~Chrysanthou, and D.~Lischinski.
\newblock Crowds by example.
\newblock In \emph{Computer graphics forum}, 2007.

\bibitem[Helbing and Molnar(1995)]{helbing1995social}
D.~Helbing and P.~Molnar.
\newblock Social force model for pedestrian dynamics.
\newblock \emph{Physical review E}, 1995.

\bibitem[Yamaguchi et~al.(2011)Yamaguchi, Berg, Ortiz, and
  Berg]{yamaguchi2011you}
K.~Yamaguchi, A.~C. Berg, L.~E. Ortiz, and T.~L. Berg.
\newblock Who are you with and where are you going?
\newblock In \emph{CVPR}, 2011.

\bibitem[Mangalam et~al.(2020)Mangalam, Girase, Agarwal, Lee, Adeli, Malik, and
  Gaidon]{mangalam2020not}
K.~Mangalam, H.~Girase, S.~Agarwal, K.-H. Lee, E.~Adeli, J.~Malik, and
  A.~Gaidon.
\newblock It is not the journey but the destination: Endpoint conditioned
  trajectory prediction.
\newblock In \emph{ECCV}, 2020.

\bibitem[Dwivedi et~al.(2020)Dwivedi, Malla, Dariush, and Choi]{dwivedi2020ssp}
I.~Dwivedi, S.~Malla, B.~Dariush, and C.~Choi.
\newblock Ssp: Single shot future trajectory prediction.
\newblock In \emph{IROS}, 2020.

\bibitem[Lee et~al.(2017)Lee, Choi, Vernaza, Choy, Torr, and
  Chandraker]{lee2017desire}
N.~Lee, W.~Choi, P.~Vernaza, C.~B. Choy, P.~H. Torr, and M.~Chandraker.
\newblock Desire: Distant future prediction in dynamic scenes with interacting
  agents.
\newblock In \emph{CVPR}, pages 336--345, 2017.

\bibitem[Xue et~al.(2018)Xue, Huynh, and Reynolds]{xue2018ss}
H.~Xue, D.~Q. Huynh, and M.~Reynolds.
\newblock Ss-lstm: A hierarchical lstm model for pedestrian trajectory
  prediction.
\newblock In \emph{WACV}, 2018.

\bibitem[Tang and Salakhutdinov(2019)]{tang2019multiple}
C.~Tang and R.~R. Salakhutdinov.
\newblock Multiple futures prediction.
\newblock In \emph{NeurIPS}, 2019.

\bibitem[Salzmann et~al.(2020)Salzmann, Ivanovic, Chakravarty, and
  Pavone]{salzmann2020trajectron++}
T.~Salzmann, B.~Ivanovic, P.~Chakravarty, and M.~Pavone.
\newblock Trajectron++: Multi-agent generative trajectory forecasting with
  heterogeneous data for control.
\newblock In \emph{ECCV}, 2020.

\bibitem[Zhao et~al.(2020)Zhao, Gao, Lan, Sun, Sapp, Varadarajan, Shen, Shen,
  Chai, Schmid, et~al.]{zhao2020tnt}
H.~Zhao, J.~Gao, T.~Lan, C.~Sun, B.~Sapp, B.~Varadarajan, Y.~Shen, Y.~Shen,
  Y.~Chai, C.~Schmid, et~al.
\newblock Tnt: Target-driven trajectory prediction.
\newblock \emph{arXiv preprint arXiv:2008.08294}, 2020.

\bibitem[Choi and Dariush(2019)]{choi2019looking}
C.~Choi and B.~Dariush.
\newblock Looking to relations for future trajectory forecast.
\newblock In \emph{ICCV}, 2019.

\bibitem[Rhinehart et~al.(2019)Rhinehart, McAllister, Kitani, and
  Levine]{rhinehart2019precog}
N.~Rhinehart, R.~McAllister, K.~Kitani, and S.~Levine.
\newblock Precog: Prediction conditioned on goals in visual multi-agent
  settings.
\newblock In \emph{ICCV}, 2019.

\bibitem[Chai et~al.(2020)Chai, Sapp, Bansal, and Anguelov]{chai2019multipath}
Y.~Chai, B.~Sapp, M.~Bansal, and D.~Anguelov.
\newblock Multipath: Multiple probabilistic anchor trajectory hypotheses for
  behavior prediction.
\newblock In \emph{CoRL}, 2020.

\bibitem[Li et~al.(2019{\natexlab{a}})Li, Ma, Zhan, and
  Tomizuka]{li2019coordination}
J.~Li, H.~Ma, W.~Zhan, and M.~Tomizuka.
\newblock Coordination and trajectory prediction for vehicle interactions via
  bayesian generative modeling.
\newblock In \emph{IV}, 2019{\natexlab{a}}.

\bibitem[Li et~al.(2019{\natexlab{b}})Li, Ying, and Chuah]{li2019grip}
X.~Li, X.~Ying, and M.~C. Chuah.
\newblock Grip: Graph-based interaction-aware trajectory prediction.
\newblock In \emph{ITSC}, 2019{\natexlab{b}}.

\bibitem[Huang et~al.(2018)Huang, Cheng, Geng, Cao, Zhou, Wang, Lin, and
  Yang]{huang2018apolloscape}
X.~Huang, X.~Cheng, Q.~Geng, B.~Cao, D.~Zhou, P.~Wang, Y.~Lin, and R.~Yang.
\newblock The apolloscape dataset for autonomous driving.
\newblock In \emph{CVPRW}, 2018.

\bibitem[Caesar et~al.(2020)Caesar, Bankiti, Lang, Vora, Liong, Xu, Krishnan,
  Pan, Baldan, and Beijbom]{nuscenes2019}
H.~Caesar, V.~Bankiti, A.~H. Lang, S.~Vora, V.~E. Liong, Q.~Xu, A.~Krishnan,
  Y.~Pan, G.~Baldan, and O.~Beijbom.
\newblock nuscenes: A multimodal dataset for autonomous driving.
\newblock In \emph{CVPR}, 2020.

\bibitem[Sun et~al.(2020)Sun, Kretzschmar, Dotiwalla, Chouard, Patnaik, Tsui,
  Guo, Zhou, Chai, Caine, et~al.]{sun2019scalability}
P.~Sun, H.~Kretzschmar, X.~Dotiwalla, A.~Chouard, V.~Patnaik, P.~Tsui, J.~Guo,
  Y.~Zhou, Y.~Chai, B.~Caine, et~al.
\newblock Scalability in perception for autonomous driving: Waymo open dataset.
\newblock In \emph{CVPR}, 2020.

\bibitem[Sch{\"o}ller et~al.(2020)Sch{\"o}ller, Aravantinos, Lay, and
  Knoll]{scholler2019simpler}
C.~Sch{\"o}ller, V.~Aravantinos, F.~Lay, and A.~Knoll.
\newblock What the constant velocity model can teach us about pedestrian motion
  prediction.
\newblock \emph{RAL}, 2020.

\bibitem[Messaoud et~al.(2019)Messaoud, Yahiaoui, Verroust-Blondet, and
  Nashashibi]{messaoud2019non}
K.~Messaoud, I.~Yahiaoui, A.~Verroust-Blondet, and F.~Nashashibi.
\newblock Non-local social pooling for vehicle trajectory prediction.
\newblock In \emph{IV}, 2019.

\bibitem[Roy et~al.(2019)Roy, Ishizaka, Mohan, and Fukuda]{roy2019vehicle}
D.~Roy, T.~Ishizaka, C.~K. Mohan, and A.~Fukuda.
\newblock Vehicle trajectory prediction at intersections using interaction
  based generative adversarial networks.
\newblock In \emph{ITSC}, 2019.

\bibitem[Bi et~al.(2019)Bi, Fang, Mao, Wang, and Deng]{bi2019joint}
H.~Bi, Z.~Fang, T.~Mao, Z.~Wang, and Z.~Deng.
\newblock Joint prediction for kinematic trajectories in
  vehicle-pedestrian-mixed scenes.
\newblock In \emph{CVPR}, 2019.

\bibitem[Sadeghian et~al.(2018)Sadeghian, Legros, Voisin, Vesel, Alahi, and
  Savarese]{sadeghian2018car}
A.~Sadeghian, F.~Legros, M.~Voisin, R.~Vesel, A.~Alahi, and S.~Savarese.
\newblock Car-net: Clairvoyant attentive recurrent network.
\newblock In \emph{ECCV}, 2018.

\bibitem[Sadeghian et~al.(2019)Sadeghian, Kosaraju, Sadeghian, Hirose,
  Rezatofighi, and Savarese]{sadeghian2018sophie}
A.~Sadeghian, V.~Kosaraju, A.~Sadeghian, N.~Hirose, H.~Rezatofighi, and
  S.~Savarese.
\newblock Sophie: An attentive gan for predicting paths compliant to social and
  physical constraints.
\newblock \emph{CVPR}, 2019.

\bibitem[Kosaraju et~al.(2019)Kosaraju, Sadeghian, Mart{\'\i}n-Mart{\'\i}n,
  Reid, Rezatofighi, and Savarese]{kosaraju2019social}
V.~Kosaraju, A.~Sadeghian, R.~Mart{\'\i}n-Mart{\'\i}n, I.~Reid, H.~Rezatofighi,
  and S.~Savarese.
\newblock Social-bigat: Multimodal trajectory forecasting using bicycle-gan and
  graph attention networks.
\newblock In \emph{NeurIPS}, 2019.

\bibitem[Hu et~al.(2020)Hu, Chen, Zhang, and Gu]{hu2020collaborative}
Y.~Hu, S.~Chen, Y.~Zhang, and X.~Gu.
\newblock Collaborative motion prediction via neural motion message passing.
\newblock In \emph{CVPR}, 2020.

\bibitem[Mohamed et~al.(2020)Mohamed, Qian, Elhoseiny, and
  Claudel]{mohamed2020social}
A.~Mohamed, K.~Qian, M.~Elhoseiny, and C.~Claudel.
\newblock Social-stgcnn: A social spatio-temporal graph convolutional neural
  network for human trajectory prediction.
\newblock In \emph{CVPR}, 2020.

\end{thebibliography}

\clearpage

\appendix
\section*{Supplementary Material}

\section{Contribution}
The main contributions of the proposed paper are summarized as follows: 
\begin{itemize}
    \item Propose a trajectory forecast framework to estimate the intention of vehicles by analyzing their relational behavior.
    \item Reason about the behavior of agents as trajectories conditioned on their intentional destination and intermediate configuration for more accurate prediction.
    \item Create a new vehicle trajectory dataset with highly interactive scenarios at road intersections in urban areas and residential areas.
    \item Generalize the proposed framework to the pedestrian trajectory forecast tasks. 
\end{itemize}%(i) Propose a trajectory forecast framework to estimate the intention of vehicles by analyzing their relational behavior; (ii) Reason the behavior of agents as trajectories conditioned on their intentional destination and intermediate configuration for more accurate prediction; (iii) Create a new vehicle trajectory dataset with highly interactive scenarios at road intersections in urban areas and residential areas; and (iv) Generalize the proposed framework to the pedestrian trajectory forecast tasks. 

\begin{table*}[!b]
\centering%\vspace{-0.6cm}
  \caption{Comparison of our new dataset with the driving datasets for trajectory prediction - KITTI~\cite{Geiger2012CVPR}, Apolloscape~\cite{ma2019trafficpredict}, and Argoverse~\cite{chang2019argoverse}. 
  }\label{tbl:dataset}%\vspace{-0.2cm}
\begin{tabular}{llcccc}
\hline
 & & KITTI~\cite{Geiger2012CVPR}~&~Apolloscape ~\cite{ma2019trafficpredict}&~Argoverse~\cite{chang2019argoverse}~&~Ours \\ \hline \hline
\multicolumn{2}{l}{No. of scenarios} & 50 & 103 &- & \textbf{213}     \\ \hline
\multicolumn{2}{l}{No. of frames ($\times10^3$)} & 13.1 & {90} &\textbf{192} &59.4     \\ \hline %59470
\multicolumn{2}{l}{No. of object classes} & 8 & 5& \textbf{15}&{8}  \\ \hline 
\multicolumn{2}{l}{Sampling frequency (fps)} & 10  & 2  & 10&\textbf{10} \\ \hline
\multicolumn{2}{l}{Trajectory duration ($sec$)}& flexible &  flexible & fixed (5)&\textbf{ flexible}  \\ \hline 
\multicolumn{2}{l}{No. of intersections}&- &  -&-&\textbf{255}  \\ \hline 
\multirow{6.5}{*}{\makecell{Type of~~~\\ ~labels}} 
& 3D bounding boxes & \checkmark&no &no & \Checkmark  \\
& Ego-car odometry  & \checkmark& no&no & \Checkmark  \\
& LiDAR point cloud & \checkmark&no &\checkmark & \Checkmark  \\
& 360$^\circ$ coverage& \checkmark &no&\checkmark& \Checkmark  \\ 
& Drivable area mask& no&no&\checkmark& \Checkmark \\
& Intentional goal  &no &no&no& \Checkmark  \\
\hline
                              
\end{tabular}%\vspace{-0.5cm}
\end{table*}

\section{Honda Intersection Dataset}
% %
% A large dataset is collected in 4 urban areas in the San Francisco Bay Area. 
% %
% [[To rewrite]] This dataset includes scenarios from the Honda Research Institute 3D Dataset (H3D)~\cite{patil2019h3d} and additional scenarios annotated similarly as those in~\cite{patil2019h3d}. 
% %
% About 200 highly interactive scenes at complex intersections are chosen. 
% %
% It consists of LiDAR-based point clouds, 3D bounding boxes of traffic participants (8 classes including cars and pedestrians), odometry of the ego-car, and the corresponding 'zone' for each traffic participant. 
% %
% A comparison of KITTI, TrafficPredict and our dataset is shown in Table~\ref{tbl:dataset}.

% %%

% The point cloud data is acquired using a Velodyne HDL-64E S3 sensor and point cloud distortion correction is performed using the high-frequency GPS data. 
% %
% As highlighted in~\cite{patil2019h3d}, the 3D bounding boxes are labelled manually at 2Hz and linearly propagated to generate data at 10Hz. 
% %
% It comprises of the traffic participant's track-ID, 3D position (m), dimensions (m) and heading angle (rad). Odometry of the ego-car is obtained via Normal Distributive Transform (NDT)-based point cloud registration. 
% %
% The registered point cloud data is used to generate 5 intersection zones which are labelled from 0 through 4 in a clockwise direction, with 0 being the 'middle' zone~\ref{fig:main}. 
% %
% These spatial attributes are assigned to each traffic participant (forming its "intention category"). 
% %
% The dataset is available at http://usa.honda-ri.com/h3ds/
% %

A large-scale dataset is collected in the San Francisco Bay Area (San Fransisco, Mountain View, San Mateo, and Santa Cruz), focusing on highly interactive scenarios at four-way intersections. We chose 213 scenarios in both urban and residential areas, which contain interactions between road users toward an environment. Our intersection dataset consists of LiDAR-based point clouds (full 360$^\circ$ coverage), track-IDs of traffic participants, their 3D bounding boxes, object classes (8 categories including cars and pedestrians), odometry of the ego-car, heading angle (in $rad$), drivable area mask, and potential destination as zone (intentional goal). 

The point cloud data is acquired using a Velodyne HDL-64E S3 sensor, and distortion correction is performed using the high-frequency GPS data. Odometry of the ego-vehicle is obtained via NDT-based point cloud registration. The labels are manually annotated at 2Hz and linearly interpolated to generate labels at 10Hz. %, similar to those of H3D dataset~\cite{patil2019h3d}. 
We further use the registered point cloud data and divide the intersection by five regions (\textit{i.e.}, in a clockwise direction as illustrated in Fig.~\ref{fig:main}. Individual road agents are then assigned $g\in\{1,...,G\}$ to indicate which zone the agent belongs to, with respect to the ego-vehicle.

% \textcolor{red}{Fig.~\ref{fig:9}} displays several examples of the intersection, and top-down view maps generated from each scenario are visualized in \textcolor{red}{Fig.~\ref{fig:data2}-\ref{fig:data5}}. 

\begin{figure*}[!t]
  \centering%\vspace{-0.5cm}
  \begin{subfigure}[b]{0.22\linewidth}
    \includegraphics[height=0.215\textheight]{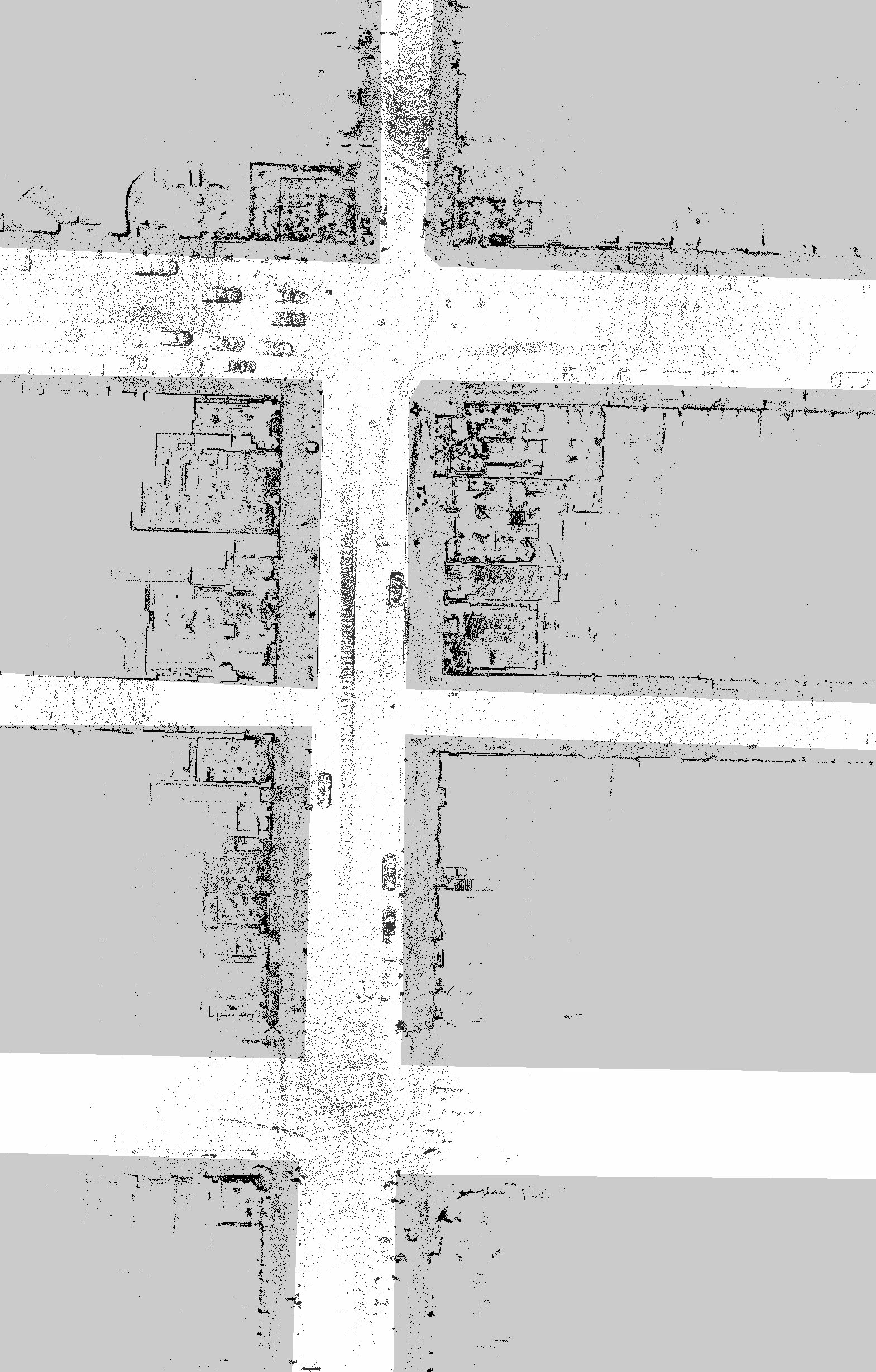}
     \caption{}\label{fig:map1}
  \end{subfigure}\quad
  \begin{subfigure}[b]{0.22\linewidth}
    \includegraphics[height=0.215\textheight]{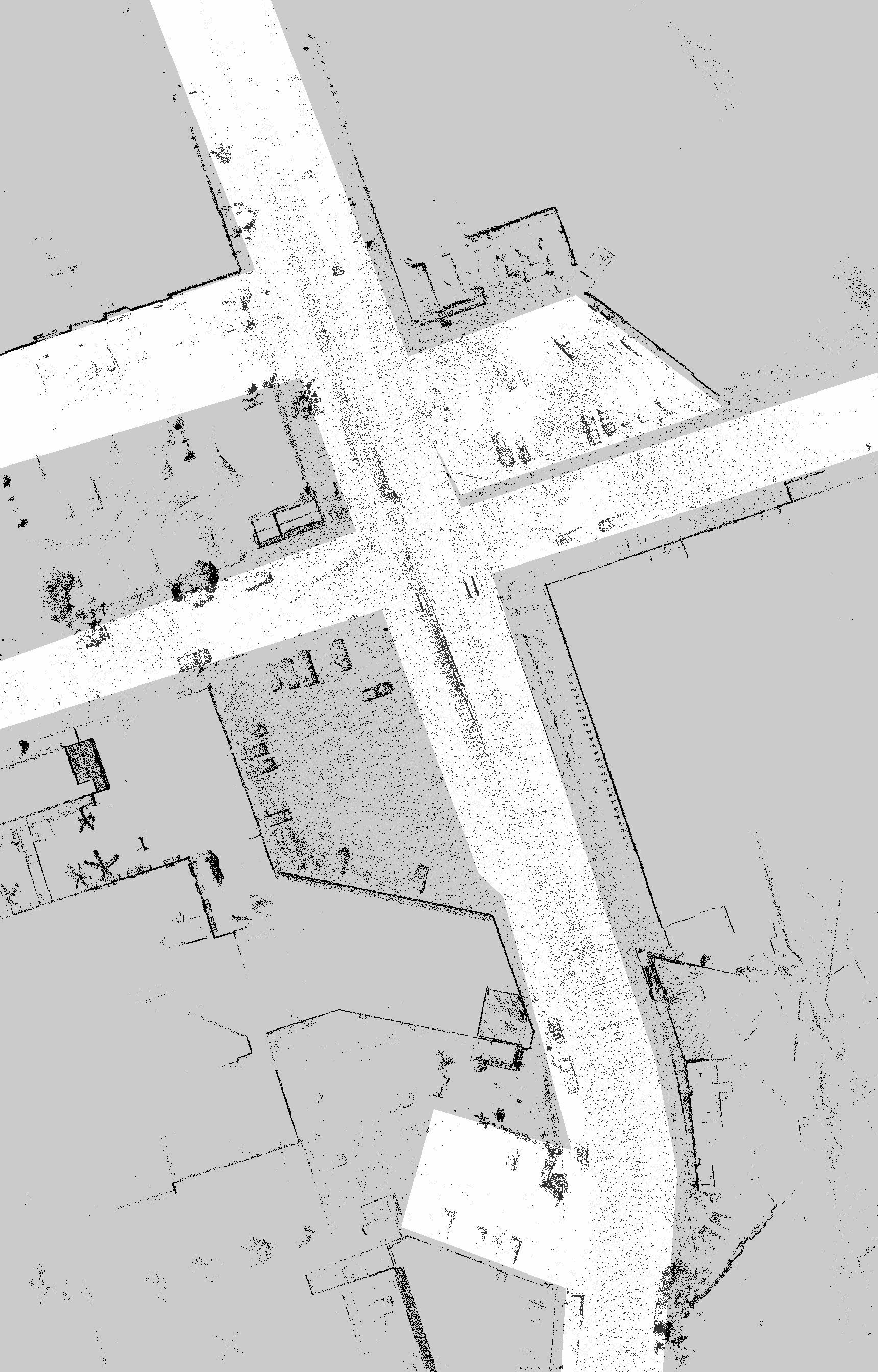}
    \caption{}\label{fig:map2}
  \end{subfigure}\quad
  \begin{subfigure}[b]{0.26\linewidth}
    \includegraphics[height=0.215\textheight]{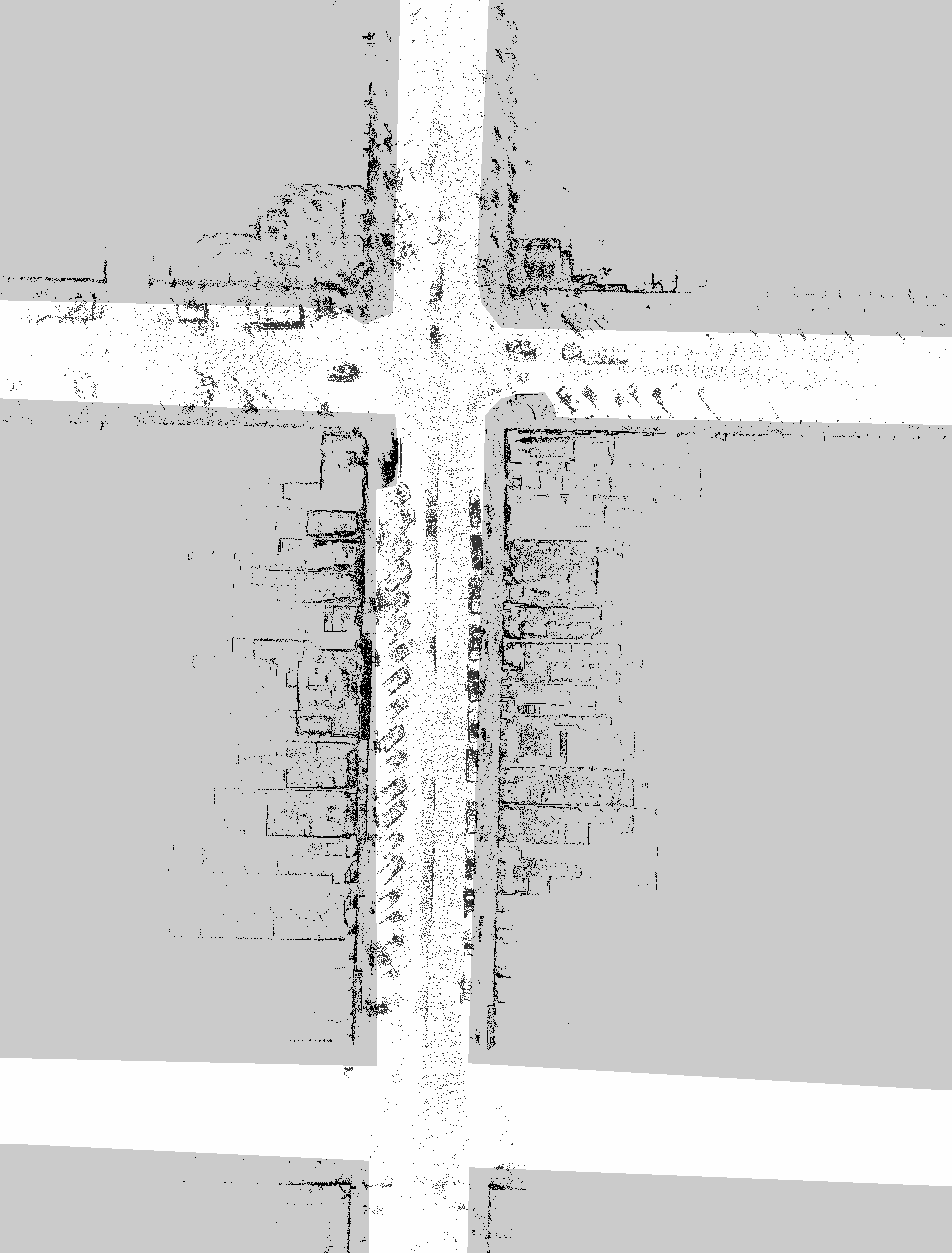}
    \caption{}\label{fig:map3}
  \end{subfigure}\quad
  \begin{subfigure}[b]{0.2\linewidth}
    \includegraphics[height=0.215\textheight]{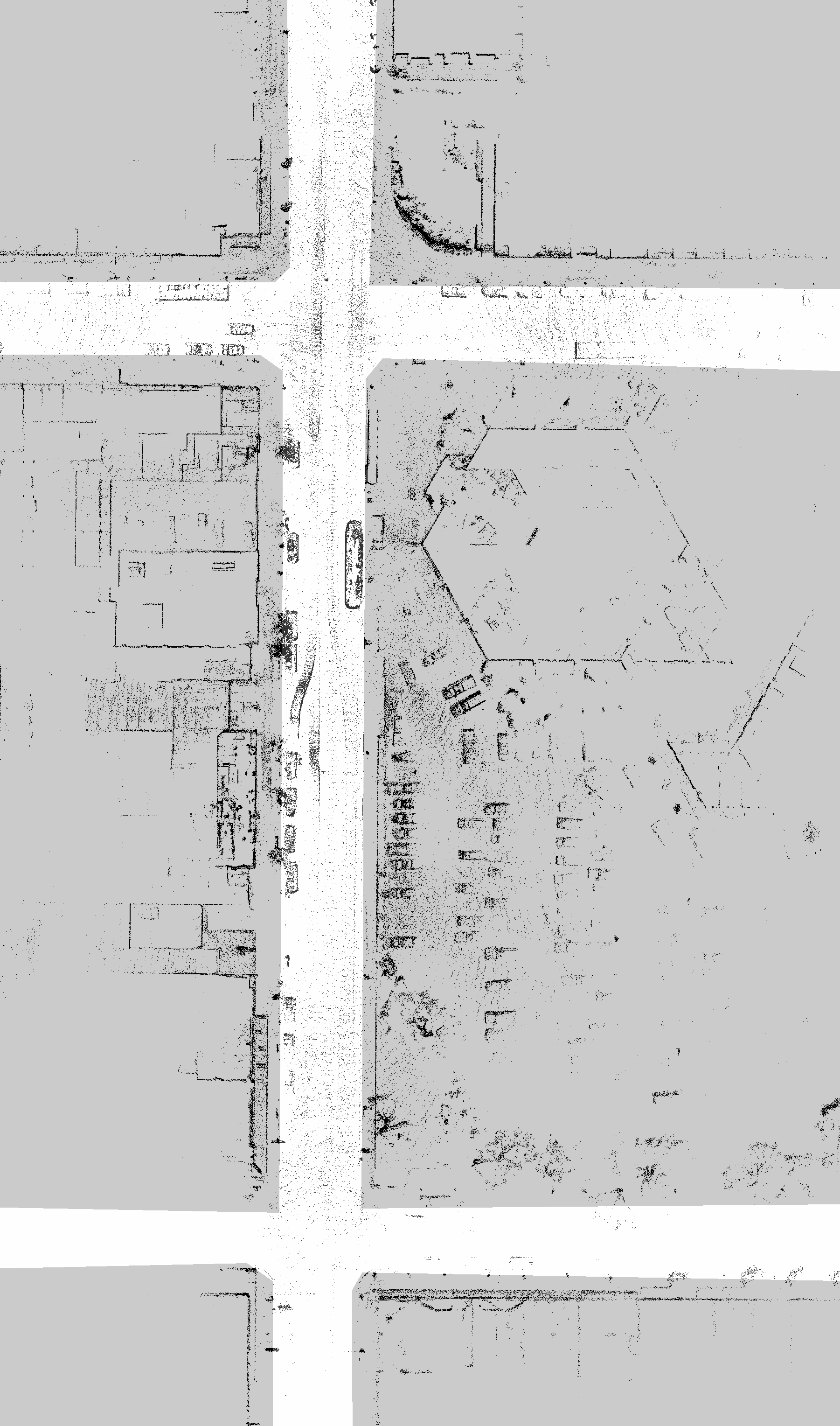}
    \caption{}\label{fig:map4}
  \end{subfigure}\\
  \begin{subfigure}[b]{0.5\linewidth}
    \includegraphics[height=0.214\textheight]{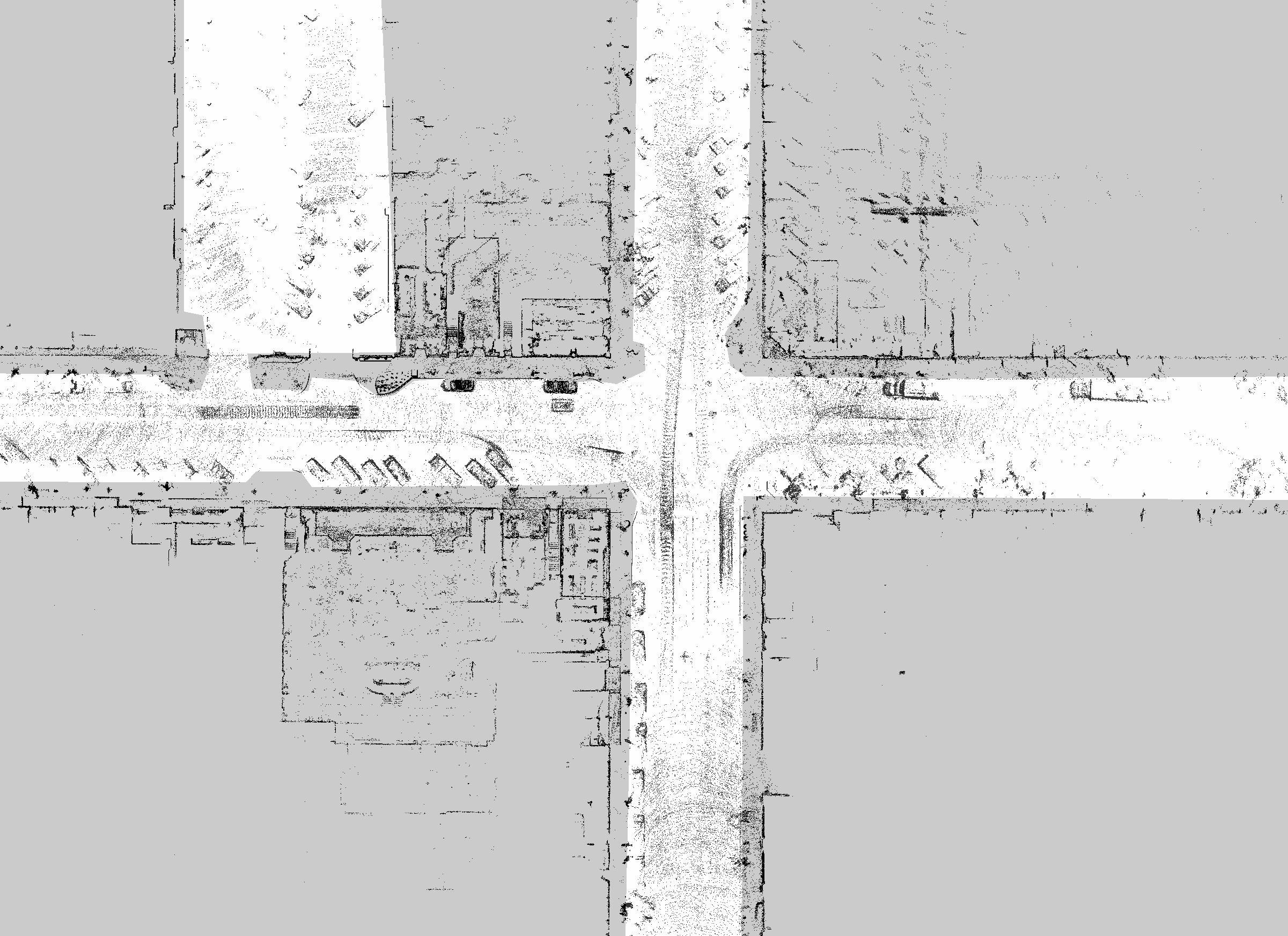}
    \caption{}\label{fig:map5}
  \end{subfigure}
  \begin{subfigure}[b]{0.48\linewidth}
    \includegraphics[height=0.214\textheight]{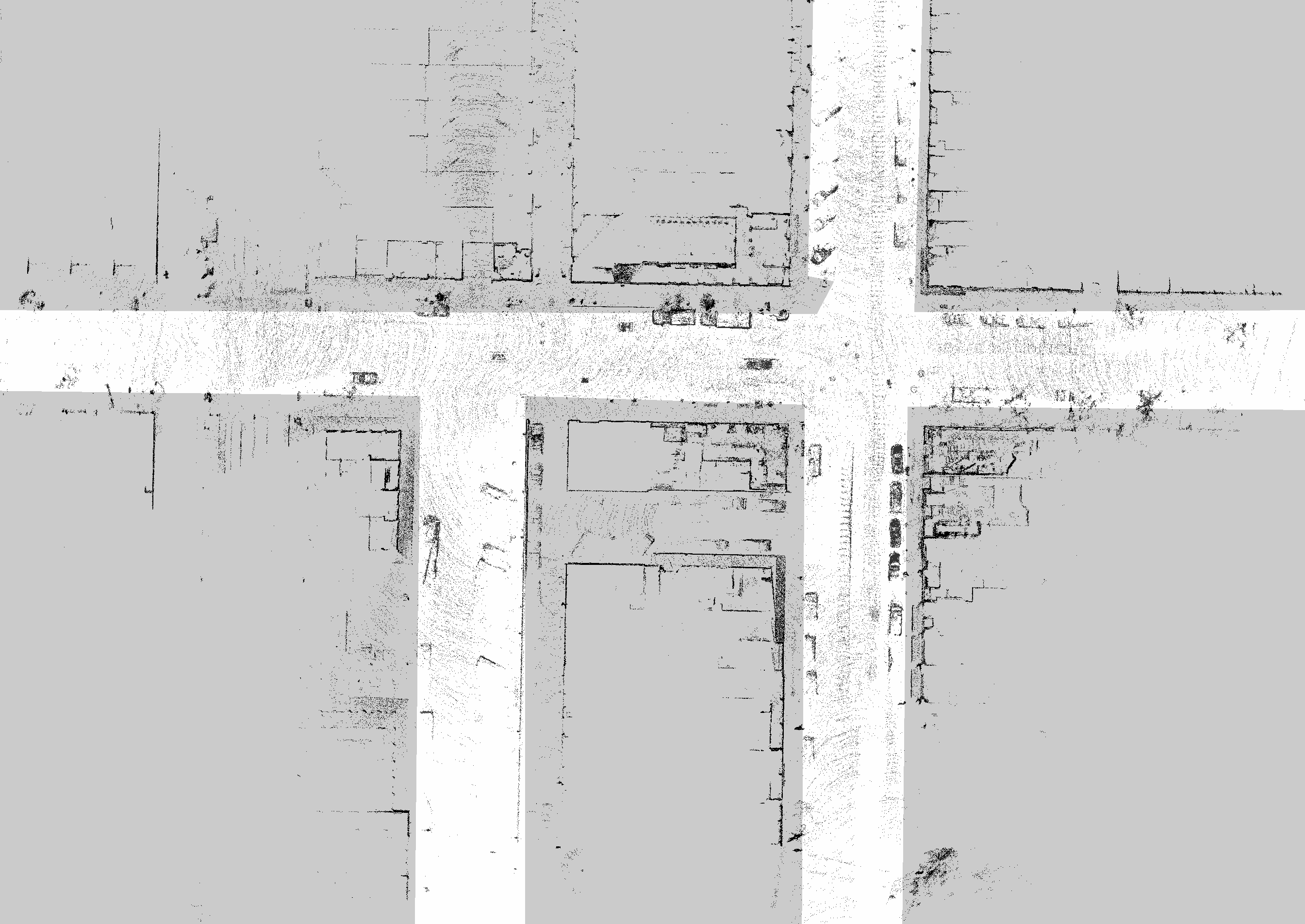}
    \caption{}\label{fig:map6}
  \end{subfigure}\\
\hspace{-0.5cm}  \begin{subfigure}[b]{0.18\linewidth}
    \includegraphics[height=0.192\textheight]{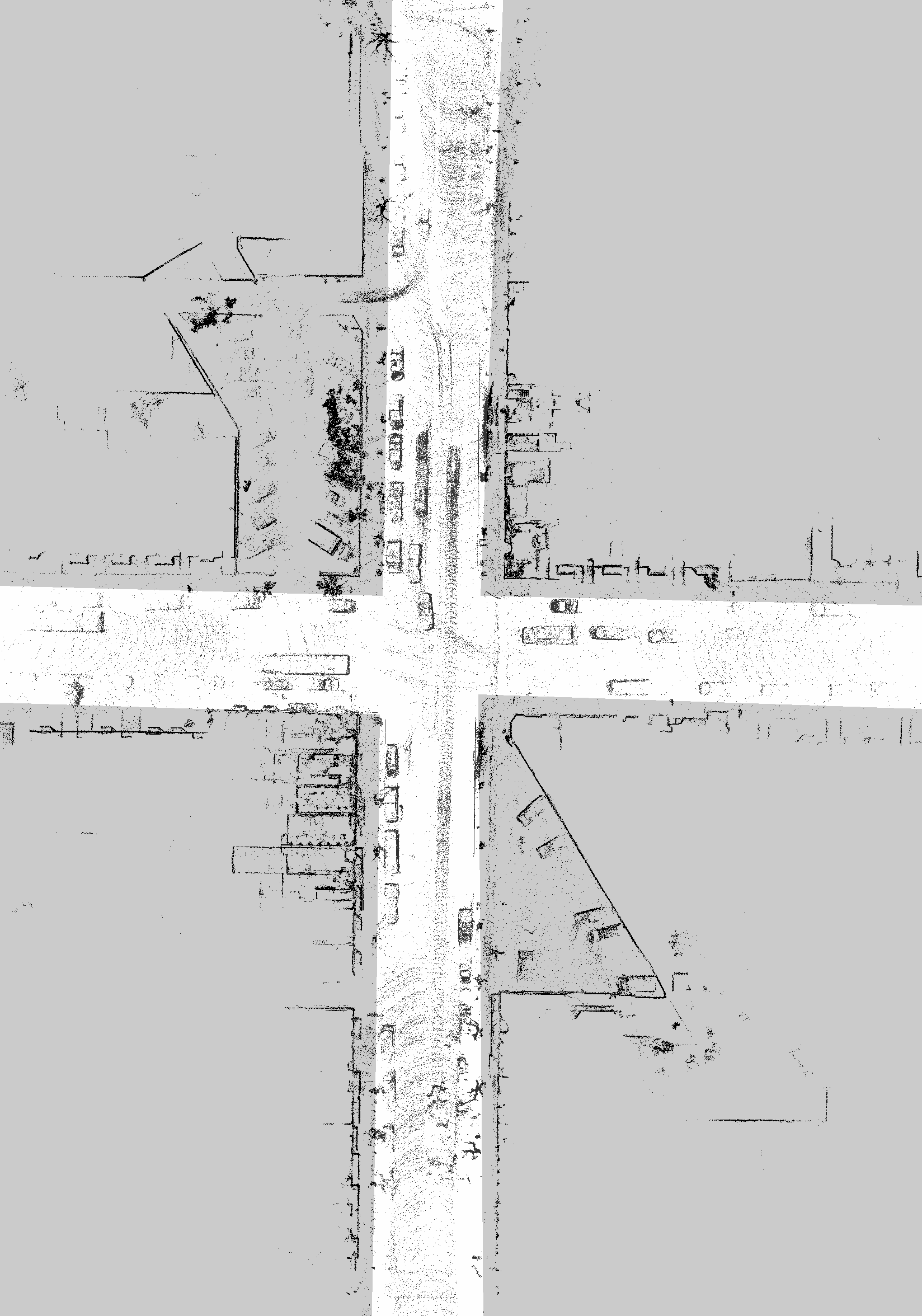}
    \caption{}\label{fig:map7}
  \end{subfigure}\quad
\hspace{0.3cm}  \begin{subfigure}[b]{0.15\linewidth}
    \includegraphics[height=0.192\textheight]{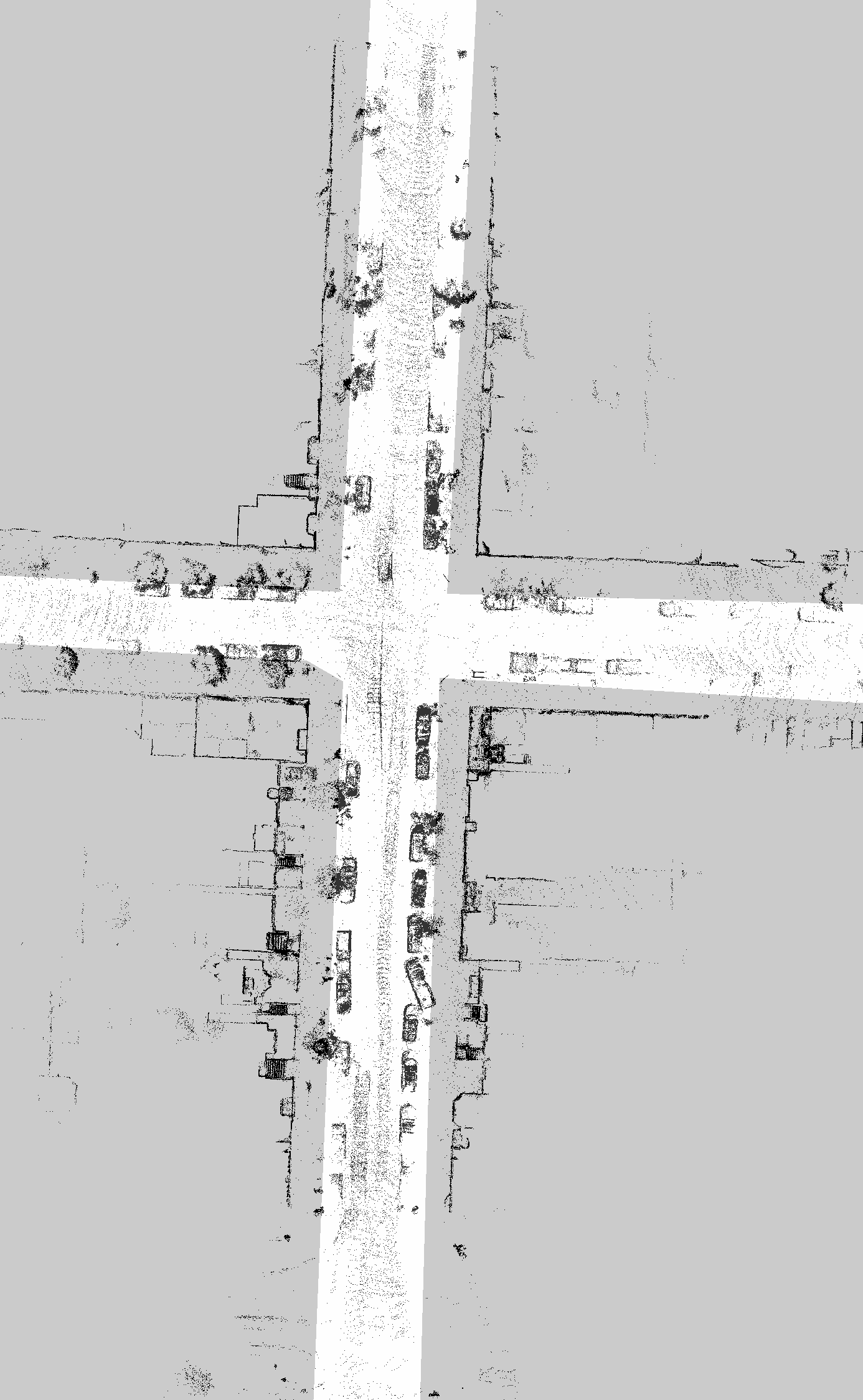}
    \caption{}\label{fig:map8}
  \end{subfigure}\quad
\hspace{0.3cm}  \begin{subfigure}[b]{0.22\linewidth}
    \includegraphics[height=0.192\textheight]{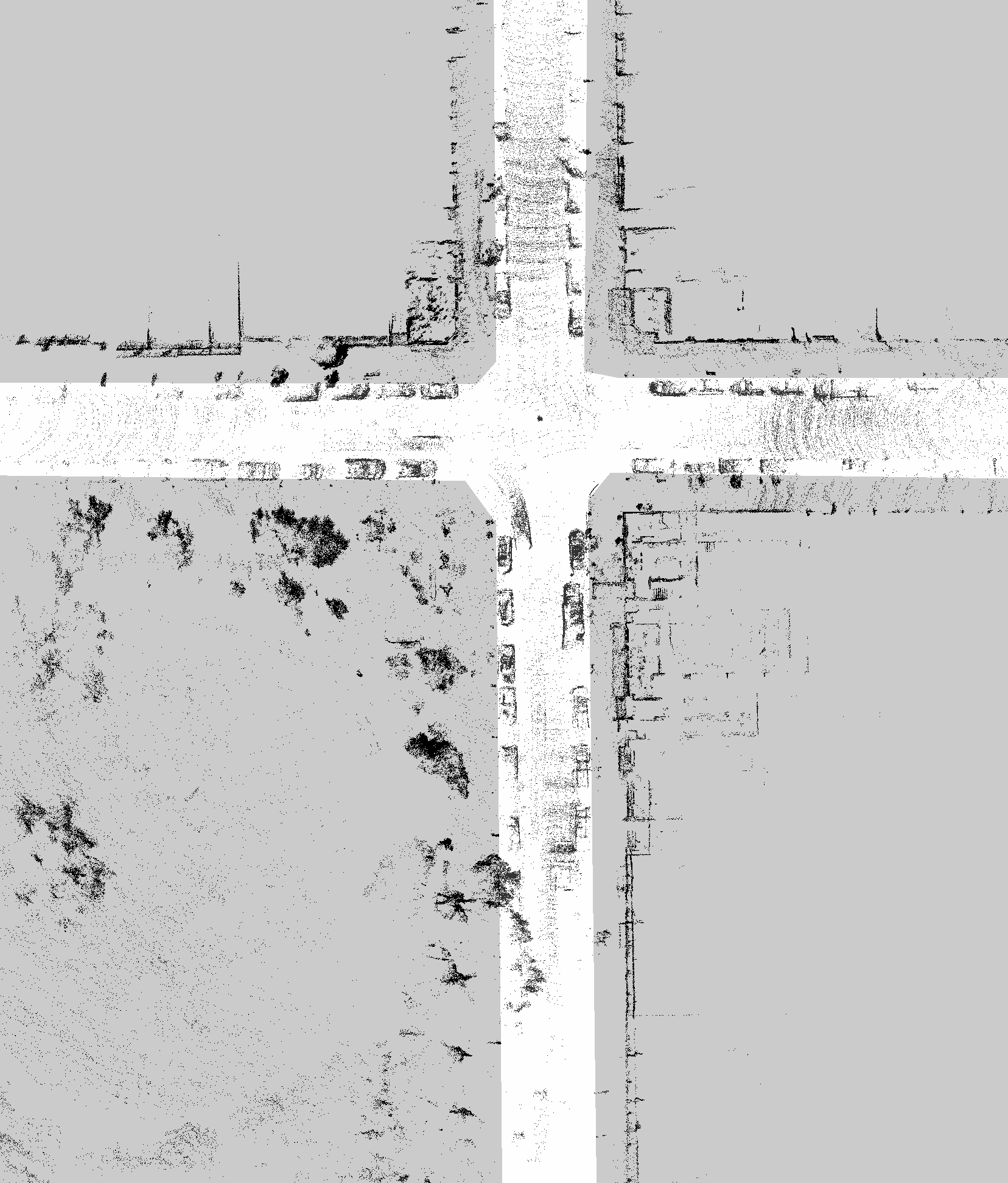}
    \caption{}\label{fig:map9}
  \end{subfigure}\quad
\hspace{0.4cm}   \begin{subfigure}[b]{0.28\linewidth}
    \includegraphics[height=0.192\textheight]{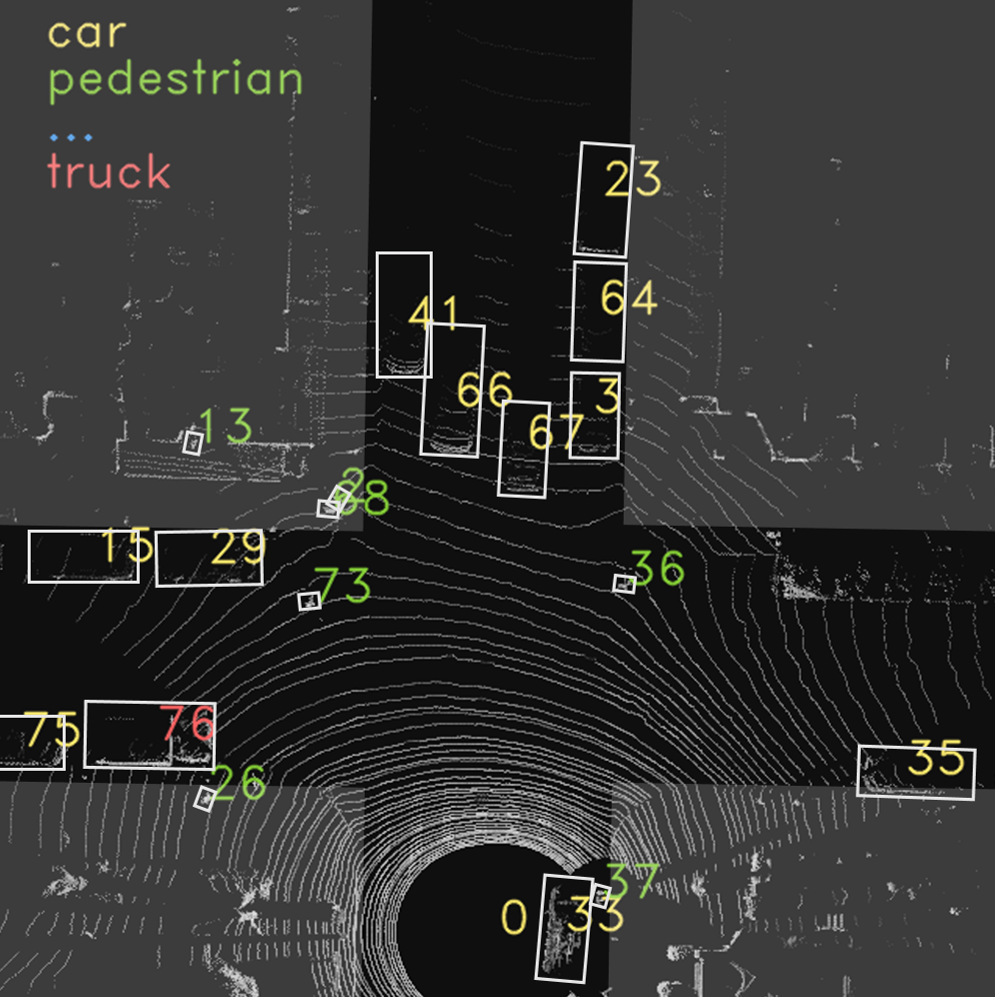}
    \caption{}\label{fig:map10}
  \end{subfigure}%\vspace{-0.2cm}
   \caption{Bird-eye view maps of interaction scenarios created using our intersection dataset. Gray mask is shown for non-drivable region.}
  \label{fig:9}
\end{figure*}

In Fig.~\ref{fig:9}, we visualize 10 example scenarios. (a-i) are the bird-eye view maps we created using LiDAR point clouds. Note that the gray mask displays non-drivable regions. Each scenario mainly focuses on interactions at the four-way intersection, and some cases additionally include other types of roads such as three-way intersection (b,f), four-way intersection (a,c,d), or parking lot (b,e). Fig.~\ref{fig:9} (g-i) shows four-way intersections in urban areas and residential areas. The preprocessed example is visualized in (j). Also, we compare our intersection dataset with the existing trajectory datasets~\cite{Geiger2012CVPR,ma2019trafficpredict,chang2019argoverse} in Table~\ref{tbl:dataset} and highlight unique features of the dataset. %Our dataset will be made available with the acceptance of this paper: \href{url}{http://}

% \begin{figure}[h]
% \begin{center}
%   \mbox{
%     \subfigure[ \label{fig:map1}]{\includegraphics[height=0.215\textheight]{figures/10.png}}\quad
%     \subfigure[ \label{fig:map2}]{\includegraphics[height=0.215\textheight]{figures/11.png}}\quad
%     \subfigure[ \label{fig:map3}]{\includegraphics[height=0.215\textheight]{figures/12.png}}\quad
%     \subfigure[ \label{fig:map4}]{\includegraphics[height=0.215\textheight]{figures/8.png}}
%     }
%   \mbox{
%     \subfigure[ \label{fig:map5}]{\includegraphics[height=0.214\textheight]{figures/14.png}}\quad
%     \subfigure[ \label{fig:map6}]{\includegraphics[height=0.214\textheight]{figures/15.png}}}
% \mbox{
%     \subfigure[ \label{fig:map7}]{\includegraphics[height=0.192\textheight]{figures/5.png}}\quad
%     \subfigure[ \label{fig:map8}]{\includegraphics[height=0.192\textheight]{figures/6.png}}\quad
%     \subfigure[ \label{fig:map9}]{\includegraphics[height=0.192\textheight]{figures/9.png}}\quad
%     \subfigure[ \label{fig:map10}]{\includegraphics[height=0.192\textheight]{figures/17.png}}
%     }
%   \caption{Bird-eye view maps of interaction scenarios created using our intersection dataset. Gray mask is shown for non-drivable region.}
%   \label{fig:9}
% \end{center}
% \end{figure}

\vspace{2.2cm}
\section{Additional Evaluation}
\subsection{Qualitative Results}

We conduct additional qualitative evaluation using the presented intersection dataset.

\begin{figure*}[!h]
\centering%\vspace{-0.2cm}
\begin{subfigure}[b]{0.17\linewidth}
    \includegraphics[width=\linewidth]{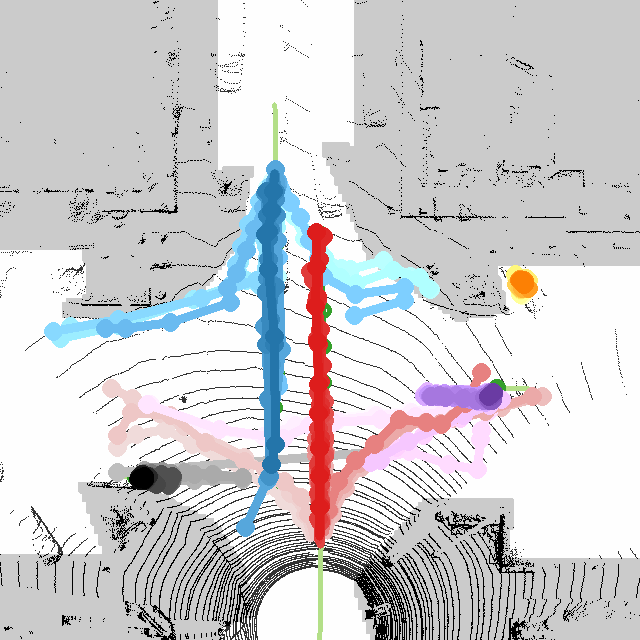}
     \caption{}%\label{fig:sta1}
  \end{subfigure}\quad
  \begin{subfigure}[b]{0.17\linewidth}
    \includegraphics[width=\linewidth]{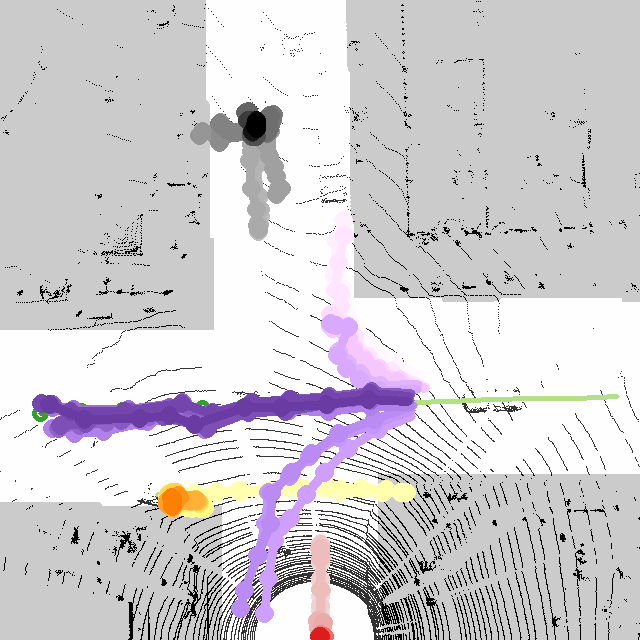}
    \caption{}%\label{fig:sta1}
  \end{subfigure}\quad
  \begin{subfigure}[b]{0.17\linewidth}
    \includegraphics[width=\linewidth]{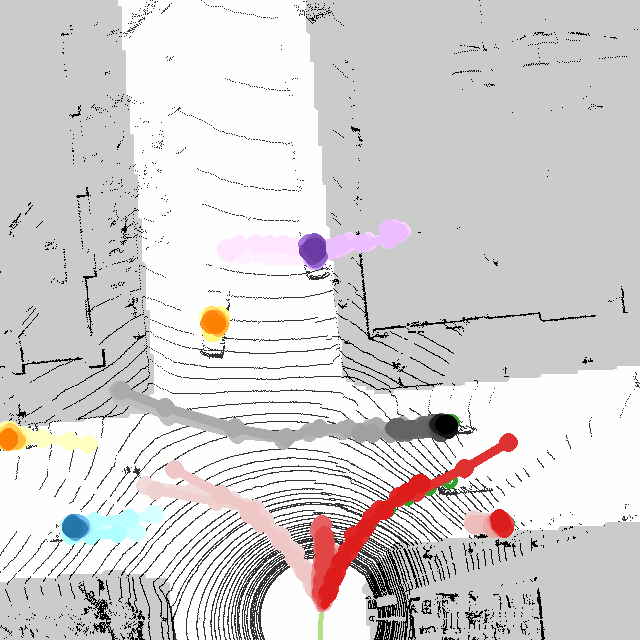}
    \caption{}%\label{fig:sta1}
  \end{subfigure}\quad
  \begin{subfigure}[b]{0.17\linewidth}
    \includegraphics[width=\linewidth]{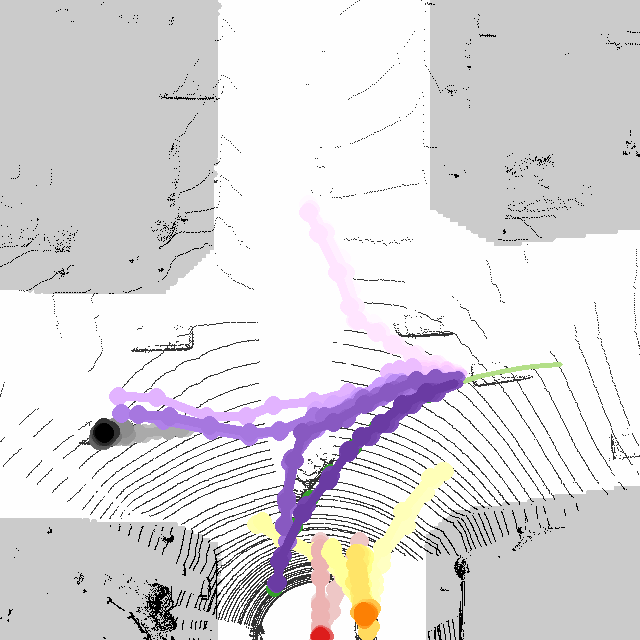}
     \caption{}%\label{fig:sta1}
  \end{subfigure}\quad
  \begin{subfigure}[b]{0.17\linewidth}
    \includegraphics[width=\linewidth]{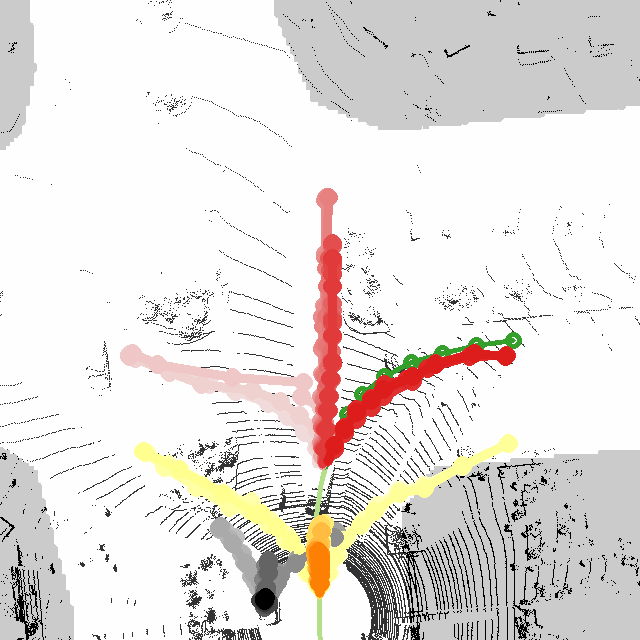}
    \caption{}%\label{fig:sta1}
  \end{subfigure}
\caption{All 20 trajectories of DROGON-Prob-20 are plotted for multiple vehicles interactions. We change the brightness of a color for those 20 samples and use different color for different vehicles. DROGON accordingly predicts goal-oriented trajectories based on the intention of vehicles. Gray mask is shown for non-drivable region.}\label{fig:5}
\end{figure*}%\vspace{-1cm}
\begin{figure*}[!h]
  \centering
  \begin{subfigure}[b]{0.25\linewidth}
    \includegraphics[width=\linewidth]{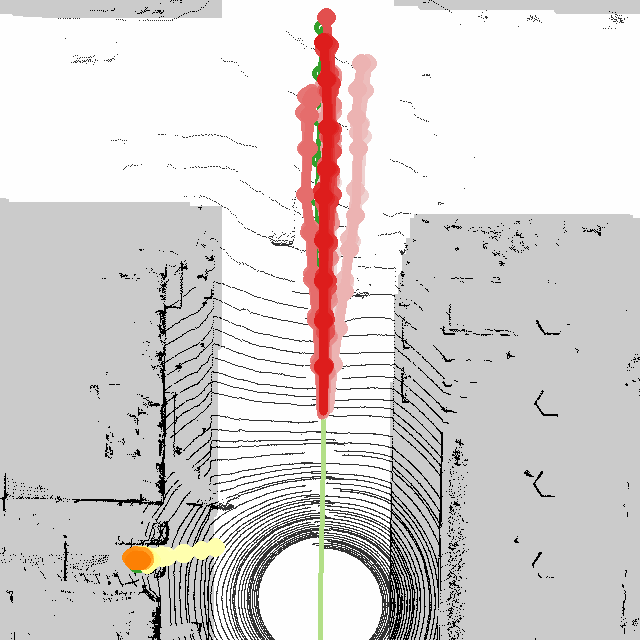}
     \caption{}\label{fig:sta1}
  \end{subfigure}\quad
  \begin{subfigure}[b]{0.25\linewidth}
    \includegraphics[width=\linewidth]{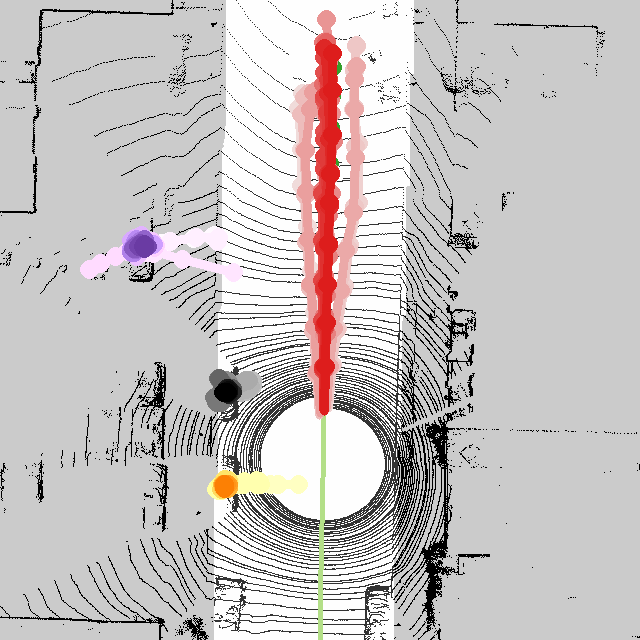}
    \caption{}\label{fig:sta1}
  \end{subfigure}\quad%\vspace{-0.2cm}
  \caption{All 20 trajectories of DROGON-Prob-20 are plotted for multiple vehicles interactions. DROGON accordingly predicts goal-oriented trajectories based on the intention of vehicles while approaching the intersection. Gray mask is shown for non-drivable region.}
  \label{fig:6}
\end{figure*}

In Fig.~\ref{fig:5}, we visualize all 20 trajectories generated from DROGON-Prob-20. By considering road layouts and interactions of each agent with others, the proposed framework appropriately generates goal-oriented trajectories conditioned on their estimated intentions. The output trajectories are inherently multi-modal. In Fig.~\ref{fig:6}, we also visualize different road layouts other than intersections. DROGON is capable of generating accurate trajectories with respect to road types.

\begin{figure*}[!t]%\vspace{0.1cm}
  \centering
  \begin{subfigure}[b]{0.17\linewidth}
    \includegraphics[width=\linewidth]{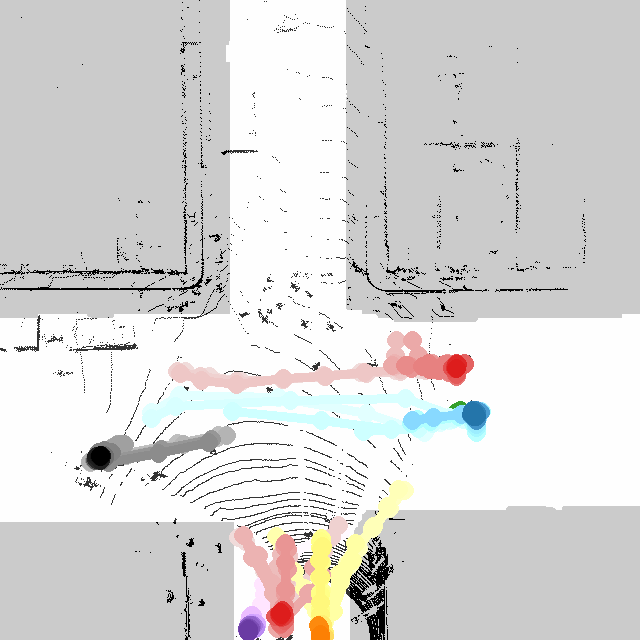}
     \caption{}\label{fig:sta1}
  \end{subfigure}\quad
  \begin{subfigure}[b]{0.17\linewidth}
    \includegraphics[width=\linewidth]{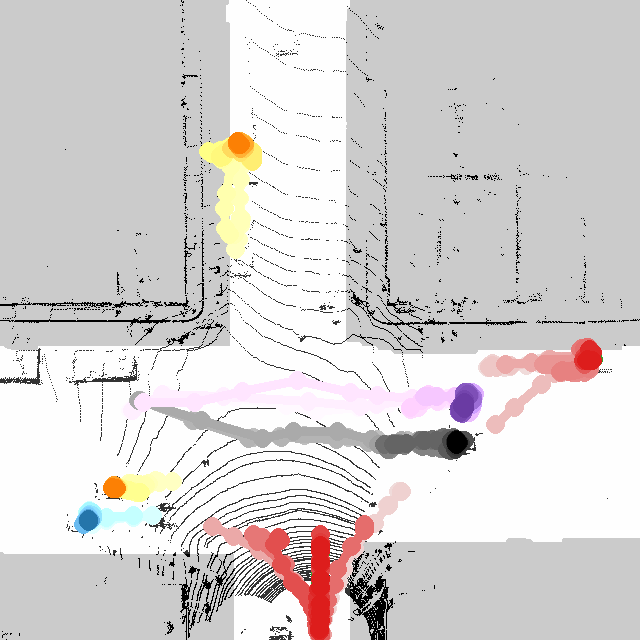}
    \caption{}\label{fig:sta1}
  \end{subfigure}\quad
  \begin{subfigure}[b]{0.17\linewidth}
    \includegraphics[width=\linewidth]{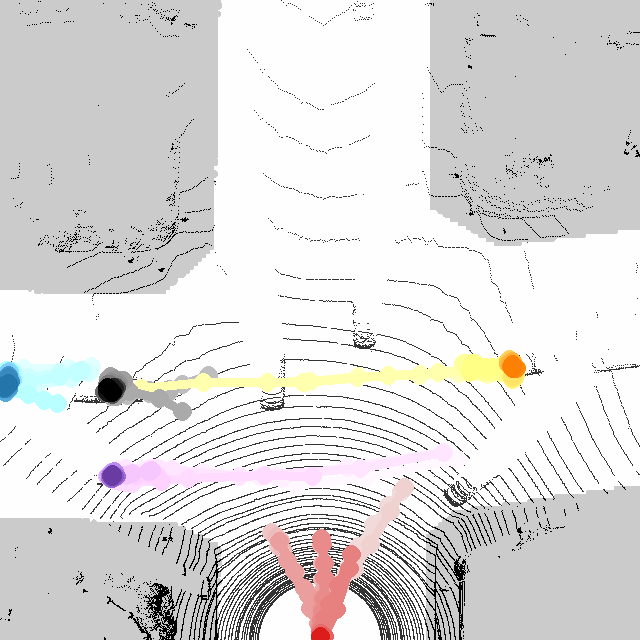}
    \caption{}\label{fig:sta1}
  \end{subfigure}\quad
  \begin{subfigure}[b]{0.17\linewidth}
    \includegraphics[width=\linewidth]{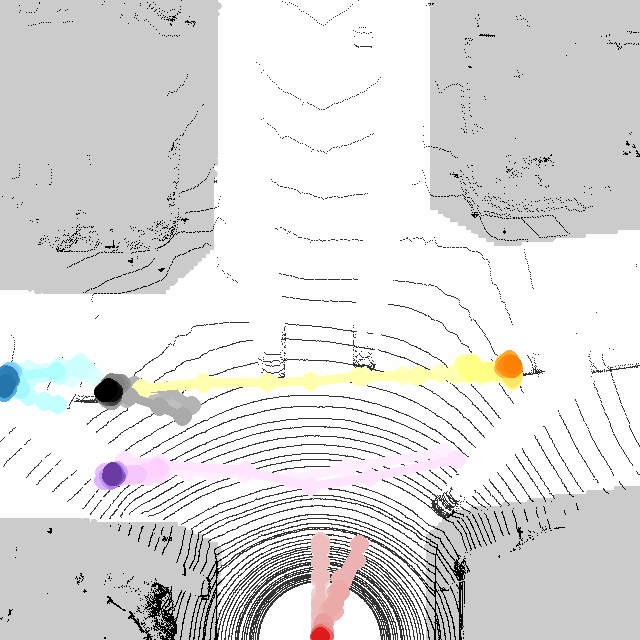}
     \caption{}\label{fig:sta1}
  \end{subfigure}\quad
  \begin{subfigure}[b]{0.17\linewidth}
    \includegraphics[width=\linewidth]{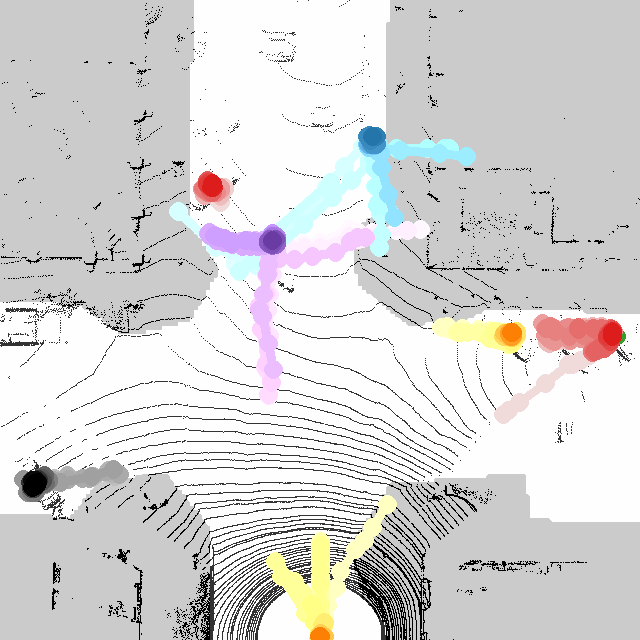}
    \caption{}\label{fig:sta1}
  \end{subfigure}
  \caption{While the vehicles are stopped, our approach is still capable of predicting their potential movements. This is apparently helpful to avoid potential collisions caused by static road agents.}
  \label{fig:7}%\vspace{-0.5cm}
\end{figure*}
\begin{figure*}[!t]
  \centering
  \begin{subfigure}[b]{0.2\linewidth}
    \includegraphics[width=\linewidth]{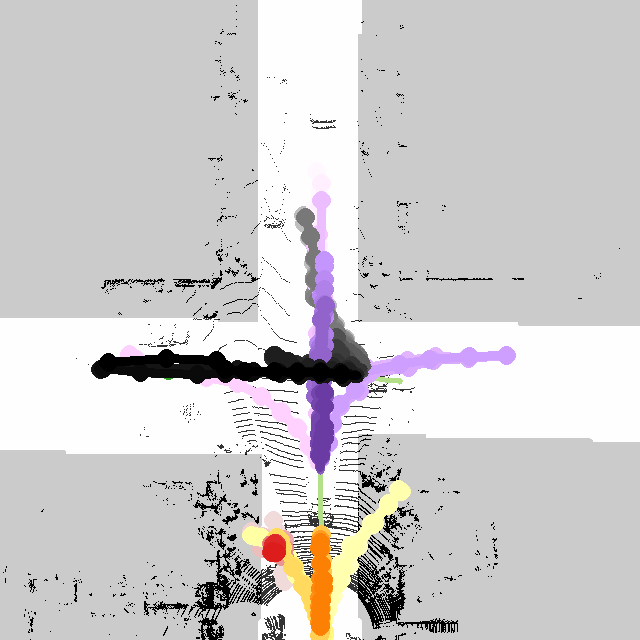}
     \caption{}\label{fig:fail1}
  \end{subfigure}\quad
  \begin{subfigure}[b]{0.2\linewidth}
    \includegraphics[width=\linewidth]{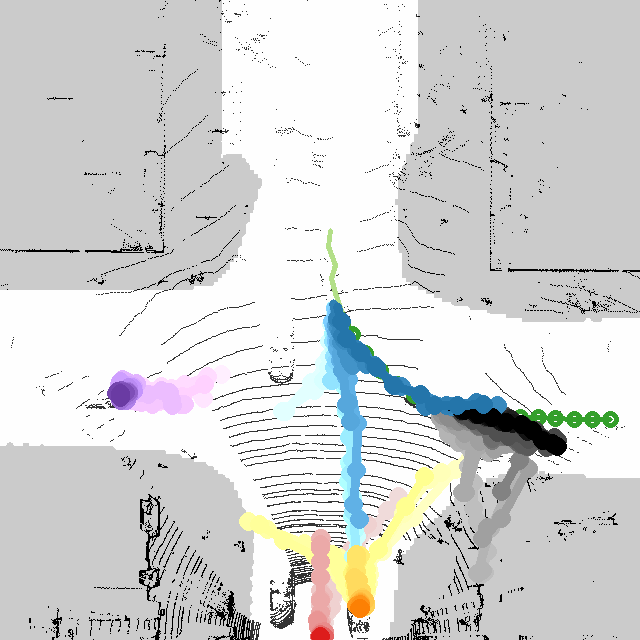}
    \caption{}\label{fig:fail2}
  \end{subfigure}\quad
  \begin{subfigure}[b]{0.2\linewidth}
    \includegraphics[width=\linewidth]{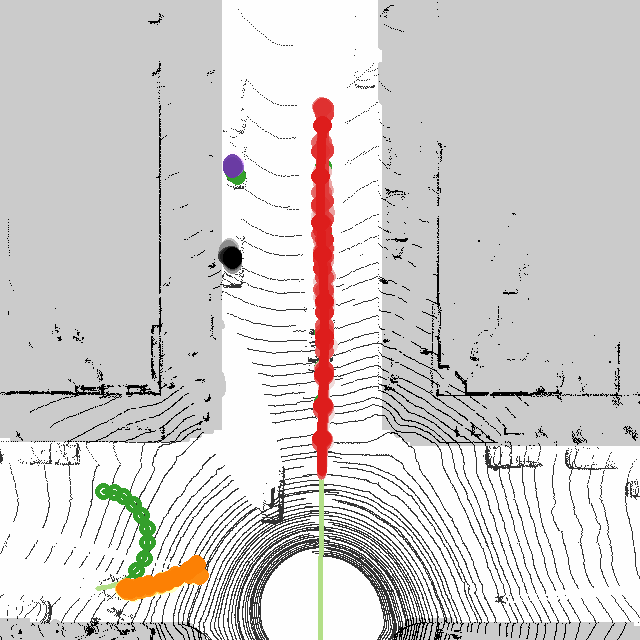}
    \caption{}\label{fig:fail3}
  \end{subfigure}
  \caption{Failure cases are visualized. (a) Sometimes, samples are drawn through non-drivable region. (b) Some predictions of on-coming vehicle (blue color) are generated toward one-way road. (c) The model experiences difficulties to predict U-turn (Dark green is the ground-truth).}
  \label{fig:8}\vspace{-0.5cm}
\end{figure*}

If a car is stopped through a red light, most of existing approaches predict its future motion as a static point based on the observation. However, our DROGON framework is able to predict their future dynamic motion given the intentional destination. By conditioning the model on the intentions, we can encourage the system to generate potential trajectories as shown in Fig.~\ref{fig:7}. This is apparently helpful to avoid potential collisions that might be caused by unexpected motion of static road agents.

We visualize failure cases in Fig.~\ref{fig:8} where (a) trajectories are generated through the non-drivable region (light yellow toward the bottom-right corner), (b) the model does not recognize one-way road (blue toward the bottom), and (c) predictions are not made for u-turn (green on the left). In this view, our future plans include (a) a design of more powerful penalty terms on top of non-drivable region masks, (b) a use of full semantics such as road signs and signals from supplemental RGB images, and (c) a collection of more diverse driving activities to cover such scenarios with more samples.

% We further visualize failure cases in \ref{fig:4g}-\ref{fig:4h}. The model does not recognize one-way road (blue toward yellow) in \ref{fig:4g}, and predictions are not made for u-turn (orange) in \ref{fig:4h}. In this view, our future plans include (a) use of full semantics such as road signs and signals from RGB images and (b) collection of rare driving activities on the road.

\begin{table*}[h]
\footnotesize
\centering
\captionof{table}{Quantitative comparison (ADE / FDE in normalized \textit{pixels}) of the proposed approach (DROGON-E) with the state-of-the-art methods -- S-LSTM~\cite{alahi2016social}, SS-LSTM~\cite{xue2018ss}, S-GAN~\cite{gupta2018social}, Gated-RN~\cite{choi2019looking}, SSP~\cite{dwivedi2020ssp} -- using the ETH~\cite{pellegrini2009you} and UCY~\cite{lerner2007crowds} dataset. }
\resizebox{0.98\textwidth}{!}
{    \begin{tabular}{l||lllll|l}
\hline
&ETH\_hotel&ETH\_eth&UCY\_univ&UCY\_zara01&UCY\_zara02&Average\\
\hline 
\hline
{S-LSTM} &0.076 / 0.125 &0.195 / 0.366&0.196 / 0.235&0.079 / 0.109&0.072 / 0.120&0.124 / 0.169 \\%,done
{SS-LSTM} &0.070 / 0.123 &0.095 / 0.235&0.081 / 0.131&0.050 / 0.084&0.054 / 0.091&0.070 / 0.133 \\%,done
%{S-GAN (SM)} &0.052 / 0.094 & 0.091 / 0.178 & 0.112 / 0.215 & 0.064 / 0.130 & 0.059 / 0.115 & 0.075 / 0.146 \\
{S-GAN} &0.046 / 0.081 & 0.087 / 0.169 & 0.108 / 0.206 & 0.062 / 0.127 & 0.058 / 0.114 & 0.072 / 0.139 \\%,done
{Gated-RN} & 0.018 / 0.033 & 0.052 / 0.100 & 0.064 / 0.127 & 0.044 / 0.086 & 0.030 / 0.059 & 0.044 / 0.086 \\%,done
{SSP} &0.018 / 0.031 &0.036 / 0.064&0.059 / 0.120&0.038 / 0.078&0.046 / 0.094&0.039 / 0.077 \\
\hline
{DROGON-E}&\textbf{0.011} / \textbf{0.018} &\textbf{0.022} / \textbf{0.031}&\textbf{0.035} / \textbf{0.059}&\textbf{0.030} / \textbf{0.054}&\textbf{0.018} / \textbf{0.031}&\textbf{0.020} / \textbf{0.033} \\
%0.011 / 0.018 | 0.022 / 0.031 | 0.035 / 0.059 | 0.030 / 0.054 | 0.018 / 0.031 | 0.020 / 0.033
\hline
\end{tabular}%\vspace{-0.2cm}
}
\label{tbl:quan222}\vspace{-0.4cm}
\end{table*}

\subsection{Quantitative Results}
Following the evaluation metric in~\cite{xue2018ss,choi2019looking}, we additionally report ADE / FDE at 4.8 $sec$ in normalized $pixels$ in Table~\ref{tbl:quan222}. $G=25$ is used as intentional goals. DROGON-E consistently outperforms the state-of-the-art methods \cite{alahi2016social,xue2018ss,gupta2018social,choi2019looking,dwivedi2020ssp}.

\section{Implementation Detail}

\subsection{Preprocessing}
\label{sec:preprocessing}
Every $\tau+\delta$ (past and future) number of point clouds, we first transform this subset to the local coordinates at time $t = t_0-\tau+1$ using GPS/IMU position estimates in the world coordinate. Then, we project these transformed point clouds onto the top-down image space that is discretized with a resolution of 0.5$m$. Each cell in projected top-down images $\mathcal{I}$ has a three-channel ($C_\mathcal{I}=3$) representation of the height, intensity, and density. The height and intensity is obtained by a laser scanner, and we choose the maximum value of the points in the cell. The density simply shows how many points belong to the cell and is computed by $\log(N+1)/\log(64)$, where $N$ is the number of points in the cell. We further normalize each channel to be in the range of $[0,1]$. From these projected top-down images $\mathcal{I}\in \mathbb{R}^{H\times W\times C_\mathcal{I}}$ where $H=W=160$, we create the 2D coordinates of past $\mathcal{X}$ and future trajectories $\mathcal{Y}$ in the local coordinates at time $t=t_0-\tau+1$. We further convert 2D coordinates to heatmaps $\mathcal{H}$, following \cite{choi2019looking}.

In addition, we remove dynamically moving agents (vehicles and pedestrians) from raw point clouds to only leave the stationary elements such as roads, sidewalks, buildings, and lanes, similar to \cite{lee2017desire}. Resulting point clouds are registered in the world coordinate and accordingly cropped to build a map $M\in\mathbb{R}^{H\times W\times C_M}$ in the local coordinates at $t=t_0-\tau+1$ (same as $I_{t_0-\tau+1}$). We observed that the density is always high when the ego-vehicle stops moving, and the height of the hilly road is not consistent when registered. Therefore, only the intensity channel (\textit{i.e.}, $C_M=1$) is used.

\subsection{Run Time}
The inference time is 0.0082 sec (in avg.) for the first prediction using Nvidia TITAN X GPU. Note that the output features of Stage 1 in Fig.~\ref{fig:main} can be shared by all other agents, which significantly reduces the run time to be 0.0063 sec (in avg.) from the next prediction in the same scene. We believe that our model is applicable to the autonomous driving system that runs in real-time.

\end{document}